\pdfoutput=1
\documentclass[letterpaper,11pt]{article}
\usepackage[margin=1in]{geometry}
\usepackage[
CJKbookmarks=true,
bookmarksnumbered=true,
bookmarksopen=true,
colorlinks=true,
citecolor=red,
linkcolor=blue,
anchorcolor=red,
urlcolor=blue
]{hyperref}

\usepackage[title]{appendix}
            
\usepackage{graphicx}
\usepackage{booktabs} 

\usepackage{natbib}
\usepackage{algorithm}
\usepackage{algorithmic}
\usepackage{amsmath,amsfonts,amssymb,amsthm}
\usepackage{hyperref}
\usepackage{color}
\usepackage{enumitem}
\usepackage{bm}
\usepackage{multirow}
\usepackage{bbm}
\usepackage{subfiles}
\usepackage{xspace}
\usepackage{caption}
\usepackage{mathtools} 

\newtheorem{theorem}{Theorem}
\newtheorem{lemma}{Lemma}
\newtheorem{corollary}{Corollary}

\newtheorem{prop}{Proposition}

\newtheorem{definition}{Definition}

\newtheorem{remark}{Remark}

\newcommand{\kNN}{$k$-NN}
\newcommand{\fkNN}{{\hat f^{\mathsf{NN}}}}

\newcommand{\diff}{{\mathrm{d}}}

\DeclarePairedDelimiterX{\inp}[2]{\langle}{\rangle}{#1, #2}
\DeclarePairedDelimiterX{\cbr}[1]{\{}{\}}{#1} 
\DeclarePairedDelimiterX{\rbr}[1]{(}{)}{#1} 
\DeclarePairedDelimiterX{\sbr}[1]{[}{]}{#1} 

\usepackage{prettyref}
\newrefformat{eq}{(\ref{#1})}
\newrefformat{chap}{Chapter~\ref{#1}}
\newrefformat{sec}{Section~\ref{#1}}
\newrefformat{alg}{Algorithm~\ref{#1}}
\newrefformat{fig}{Figure~\ref{#1}}
\newrefformat{tab}{Table~\ref{#1}}
\newrefformat{rmk}{Remark~\ref{#1}}
\newrefformat{clm}{Claim~\ref{#1}}
\newrefformat{def}{Definition~\ref{#1}}
\newrefformat{cor}{Corollary~\ref{#1}}
\newrefformat{lmm}{Lemma~\ref{#1}}
\newrefformat{prop}{Proposition~\ref{#1}}
\newrefformat{app}{Appendix~\ref{#1}}
\newrefformat{hyp}{Hypothesis~\ref{#1}}
\newrefformat{thm}{Theorem~\ref{#1}}
\newrefformat{ass}{Assumption~\ref{#1}}
\newrefformat{phase}{{Phase}~\ref{#1}}

\author{
	Pengkun Yang and Jingzhao Zhang
	\thanks{Accepted for presentation at the Conference on Learning Theory (COLT) 2025.}
	\thanks{
		P.\ Yang is with the Department of Statistics and Data Science, Tsinghua University, 
		\texttt{yangpengkun@tsinghua.edu.cn}.
		J.\ Zhang is with IIIS, Tsinghua University and Shanghai Qizhi Institute, \texttt{jingzhaoz@mail.tsinghua.edu.cn}. 
		P.~Yang is supported in part by the NSFC grant 12101353. P.\ Yang and J. Zhang is supported by National Key R\&D Program of China 2024YFA1015800 and Shanghai Qi Zhi Institute Innovation Program.		
	}
}

\title{Fast and Multiphase Rates for Nearest Neighbor Classifiers}

\begin{document}
	\newcommand{\defeq}{\mathrel{\mathop:}=}
\newcommand{\note}[1]{\textcolor{red}{#1}}
\newcommand{\jz}[1]{{\textcolor{red}{[#1]}}}

\newcommand{\ingd}{\textsc{Ingd}\xspace}

\newcommand{\vect}[1]{\ensuremath{\mathbf{#1}}}
\newcommand{\mat}[1]{\ensuremath{\mathbf{#1}}}
\newcommand{\dd}{\mathrm{d}}
\newcommand{\grad}{\nabla}
\newcommand{\hess}{\nabla^2}
\newcommand{\argmin}{\mathop{\mathrm{argmin}}}
\newcommand{\argmax}{\mathop{\mathrm{argmax}}}
\newcommand{\fracpar}[2]{\frac{\partial #1}{\partial  #2}}
\newcommand{\iprod}[2]{\langle #1, #2 \rangle}
\newcommand{\dkl}{D_{\text{KL}}}
\newcommand{\dtv}{D_{\text{TV}}}

\newcommand{\abs}[1]{\left|{#1}\right|}
\newcommand{\norm}[1]{\left\|{#1}\right\|}
\newcommand{\fnorm}[1]{\|{#1} \|_{\text{F}}}
\newcommand{\spnorm}[2]{\| {#1} \|_{\text{S}({#2})}}
\newcommand{\sigmin}{\sigma_{\min}}
\newcommand{\tr}{\mathrm{tr}}
\renewcommand{\det}{\mathrm{det}}
\newcommand{\rank}{\mathrm{rank}}
\newcommand{\logdet}{\mathrm{logdet}}
\newcommand{\trans}{^{\top}}
\newcommand{\poly}{\mathrm{poly}}
\newcommand{\polylog}{\mathrm{polylog}}
\newcommand{\st}{\mathrm{s.t.~}}
\newcommand{\proj}{\mathcal{P}}
\newcommand{\projII}{\mathcal{P}_{\parallel}}
\newcommand{\projT}{\mathcal{P}_{\perp}}
\newcommand{\floor}[1]{\lfloor {#1} \rfloor}

\newcommand{\cO}{\mathcal{O}}
\newcommand{\tlO}{\mathcal{\tilde{O}}}
\newcommand{\tlOmega}{\tilde{\Omega}}
\newcommand{\tlTheta}{\tilde{\Theta}}


\newcommand{\Z}{\mathbb{Z}}
\newcommand{\N}{\mathbb{N}}
\newcommand{\R}{\mathbb{R}}
\newcommand{\Sn}{\mathbb{S}}
\newcommand{\C}{\mathbb{C}}
\newcommand{\E}{\mathbb{E}}
\renewcommand{\Pr}{\mathbb{P}}
\newcommand{\Var}{\text{Var}}
\newcommand{\F}{\mathcal{F}}
\newcommand{\G}{\mathcal{G}}
\newcommand{\Y}{\mathcal{Y}}
\newcommand{\cX}{\mathcal{X}}
\newcommand{\cP}{\mathcal{P}}
\newcommand{\cS}{\mathcal{S}}


\newcommand{\A}{\mat{A}}
\newcommand{\B}{\mat{B}}
\newcommand{\Q}{\mat{Q}}

\newcommand{\I}{\mat{I}}
\newcommand{\D}{\mat{D}}
\newcommand{\U}{\mat{U}}
\newcommand{\V}{\mat{V}}
\newcommand{\W}{\mat{W}}
\newcommand{\X}{\mat{X}}
\newcommand{\mSigma}{\mat{\Sigma}}
\newcommand{\mLambda}{\mat{\Lambda}}
\newcommand{\e}{\vect{e}}
\renewcommand{\u}{\vect{u}}
\renewcommand{\v}{\vect{v}}
\newcommand{\w}{\vect{w}}
\newcommand{\x}{\vect{x}}
\newcommand{\y}{\vect{y}}
\newcommand{\z}{\vect{z}}
\newcommand{\g}{\vect{g}}
\newcommand{\zero}{\vect{0}}
\newcommand{\fI}{\mathfrak{I}}
\newcommand{\fS}{\mathfrak{S}}
\newcommand{\fE}{\mathfrak{E}}
\newcommand{\fF}{\mathfrak{F}}


\newcommand{\M}{\mathcal{M}}
\renewcommand{\L}{\mathcal{L}}
\renewcommand{\H}{\mathcal{H}}
\newcommand{\cN}{\mathcal{N}}
\newcommand{\cD}{\mathcal{D}}

\newcommand{\cn}{\kappa}
\newcommand{\nn}{\nonumber}

\newcommand{\reals}{{\mathbb{R}}}
\newcommand{\complex}{{\mathbb{C}}}
\newcommand{\integers}{{\mathbb{Z}}}
\newcommand{\naturals}{{\mathbb{N}}}
\newcommand{\rationals}{{\mathbb{Q}}}

\newcommand{\Indc}{\mathbf{1}}
\newcommand{\indc}[1]{{\mathbf{1}_{\left\{{#1}\right\}}}}

\newcommand{\iu}{{\mathbf{i}}}

\newcommand{\Th}{{ th}}
\newcommand{\eqdistr}{{\stackrel{ d.}{=}}}
\newcommand{\leqst}{{\stackrel{ st.}{\leq}}}
\newcommand{\geqst}{{\stackrel{ st.}{\geq}}}
\newcommand\indep{\protect\mathpalette{\protect\independenT}{\perp}}
\def\independenT#1#2{\mathrel{\rlap{$#1#2$}\mkern2mu{#1#2}}}

\newcommand{\Expect}{\mathbb{E}}
\newcommand{\expect}[1]{\mathbb{E}\left[#1\right]}
\newcommand{\Prob}{\mathbb{P}}
\newcommand{\prob}[1]{\mathbb{P}\left[#1\right]}
\newcommand{\iiddistr}{{\stackrel{\text{i.i.d.}}{\sim}}}
\newcommand{\Floor}[1]{\lfloor {#1} \rfloor}
\newcommand{\ceil}[1]{{\left\lceil {#1} \right \rceil}}
\newcommand{\Ceil}[1]{{\lceil {#1}  \rceil}}
\newcommand{\pth}[1]{\left( #1 \right)}
\newcommand{\qth}[1]{\left[ #1 \right]}
\newcommand{\sth}[1]{\left\{ #1 \right\}}
\newcommand{\Norm}[1]{\|{#1} \|}
\newcommand{\inner}[1]{\langle{#1}\rangle}
\newcommand{\nb}[1]{{\sf\color{blue}[#1]}}
\newcommand{\nbr}[1]{{\sf\color{red}[#1]}}
\newcommand{\nol}[1]{{\sf\color{cyan}[outline: #1 ]\par}}

\newcommand{\Unif}{\mathrm{Uniform}}
\newcommand{\Bern}{\mathrm{ Bern}}
\newcommand{\Binom}{\mathrm{ Binom}}
\newcommand{\Pois}{\mathrm{ Pois}}

\newcommand{\TV}{\mathsf{TV}}
\newcommand{\var}{\mathsf{var}}

\newcommand{\calA}{{\mathcal{A}}}
\newcommand{\calB}{{\mathcal{B}}}
\newcommand{\calC}{{\mathcal{C}}}
\newcommand{\calD}{{\mathcal{D}}}
\newcommand{\calE}{{\mathcal{E}}}
\newcommand{\calF}{{\mathcal{F}}}
\newcommand{\calG}{{\mathcal{G}}}
\newcommand{\calH}{{\mathcal{H}}}
\newcommand{\calI}{{\mathcal{I}}}
\newcommand{\calJ}{{\mathcal{J}}}
\newcommand{\calK}{{\mathcal{K}}}
\newcommand{\calL}{{\mathcal{L}}}
\newcommand{\calM}{{\mathcal{M}}}
\newcommand{\calN}{{\mathcal{N}}}
\newcommand{\calO}{{\mathcal{O}}}
\newcommand{\calP}{{\mathcal{P}}}
\newcommand{\calQ}{{\mathcal{Q}}}
\newcommand{\calR}{{\mathcal{R}}}
\newcommand{\calS}{{\mathcal{S}}}
\newcommand{\calT}{{\mathcal{T}}}
\newcommand{\calU}{{\mathcal{U}}}
\newcommand{\calV}{{\mathcal{V}}}
\newcommand{\calW}{{\mathcal{W}}}
\newcommand{\calX}{{\mathcal{X}}}
\newcommand{\calY}{{\mathcal{Y}}}
\newcommand{\calZ}{{\mathcal{Z}}}

\newcommand{\sfA}{{\mathsf{A}}}
\newcommand{\sfB}{{\mathsf{B}}}
\newcommand{\sfC}{{\mathsf{C}}}
\newcommand{\sfD}{{\mathsf{D}}}
\newcommand{\sfE}{{\mathsf{E}}}
\newcommand{\sfF}{{\mathsf{F}}}
\newcommand{\sfG}{{\mathsf{G}}}
\newcommand{\sfH}{{\mathsf{H}}}
\newcommand{\sfI}{{\mathsf{I}}}
\newcommand{\sfJ}{{\mathsf{J}}}
\newcommand{\sfK}{{\mathsf{K}}}
\newcommand{\sfL}{{\mathsf{L}}}
\newcommand{\sfM}{{\mathsf{M}}}
\newcommand{\sfN}{{\mathsf{N}}}
\newcommand{\sfO}{{\mathsf{O}}}
\newcommand{\sfP}{{\mathsf{P}}}
\newcommand{\sfQ}{{\mathsf{Q}}}
\newcommand{\sfR}{{\mathsf{R}}}
\newcommand{\sfS}{{\mathsf{S}}}
\newcommand{\sfT}{{\mathsf{T}}}
\newcommand{\sfU}{{\mathsf{U}}}
\newcommand{\sfV}{{\mathsf{V}}}
\newcommand{\sfW}{{\mathsf{W}}}
\newcommand{\sfX}{{\mathsf{X}}}
\newcommand{\sfY}{{\mathsf{Y}}}
\newcommand{\sfZ}{{\mathsf{Z}}}

\newcommand{\sfa}{{\mathsf{a}}}
\newcommand{\sfb}{{\mathsf{b}}}
\newcommand{\sfc}{{\mathsf{c}}}
\newcommand{\sfd}{{\mathsf{d}}}
\newcommand{\sfe}{{\mathsf{e}}}
\newcommand{\sff}{{\mathsf{f}}}
\newcommand{\sfg}{{\mathsf{g}}}
\newcommand{\sfh}{{\mathsf{h}}}
\newcommand{\sfi}{{\mathsf{i}}}
\newcommand{\sfj}{{\mathsf{j}}}
\newcommand{\sfk}{{\mathsf{k}}}
\newcommand{\sfl}{{\mathsf{l}}}
\newcommand{\sfm}{{\mathsf{m}}}
\newcommand{\sfn}{{\mathsf{n}}}
\newcommand{\sfo}{{\mathsf{o}}}
\newcommand{\sfp}{{\mathsf{p}}}
\newcommand{\sfq}{{\mathsf{q}}}
\newcommand{\sfr}{{\mathsf{r}}}
\newcommand{\sfs}{{\mathsf{s}}}
\newcommand{\sft}{{\mathsf{t}}}
\newcommand{\sfu}{{\mathsf{u}}}
\newcommand{\sfv}{{\mathsf{v}}}
\newcommand{\sfw}{{\mathsf{w}}}
\newcommand{\sfx}{{\mathsf{x}}}
\newcommand{\sfy}{{\mathsf{y}}}
\newcommand{\sfz}{{\mathsf{z}}}

\newcommand{\bfA}{{\mathbf{A}}}
\newcommand{\bfB}{{\mathbf{B}}}
\newcommand{\bfC}{{\mathbf{C}}}
\newcommand{\bfD}{{\mathbf{D}}}
\newcommand{\bfE}{{\mathbf{E}}}
\newcommand{\bfF}{{\mathbf{F}}}
\newcommand{\bfG}{{\mathbf{G}}}
\newcommand{\bfH}{{\mathbf{H}}}
\newcommand{\bfI}{{\mathbf{I}}}
\newcommand{\bfJ}{{\mathbf{J}}}
\newcommand{\bfK}{{\mathbf{K}}}
\newcommand{\bfL}{{\mathbf{L}}}
\newcommand{\bfM}{{\mathbf{M}}}
\newcommand{\bfN}{{\mathbf{N}}}
\newcommand{\bfO}{{\mathbf{O}}}
\newcommand{\bfP}{{\mathbf{P}}}
\newcommand{\bfQ}{{\mathbf{Q}}}
\newcommand{\bfR}{{\mathbf{R}}}
\newcommand{\bfS}{{\mathbf{S}}}
\newcommand{\bfT}{{\mathbf{T}}}
\newcommand{\bfU}{{\mathbf{U}}}
\newcommand{\bfV}{{\mathbf{V}}}
\newcommand{\bfW}{{\mathbf{W}}}
\newcommand{\bfX}{{\mathbf{X}}}
\newcommand{\bfY}{{\mathbf{Y}}}
\newcommand{\bfZ}{{\mathbf{Z}}}

\newcommand{\bfa}{{\mathbf{a}}}
\newcommand{\bfb}{{\mathbf{b}}}
\newcommand{\bfc}{{\mathbf{c}}}
\newcommand{\bfd}{{\mathbf{d}}}
\newcommand{\bfe}{{\mathbf{e}}}
\newcommand{\bff}{{\mathbf{f}}}
\newcommand{\bfg}{{\mathbf{g}}}
\newcommand{\bfh}{{\mathbf{h}}}
\newcommand{\bfi}{{\mathbf{i}}}
\newcommand{\bfj}{{\mathbf{j}}}
\newcommand{\bfk}{{\mathbf{k}}}
\newcommand{\bfl}{{\mathbf{l}}}
\newcommand{\bfm}{{\mathbf{m}}}
\newcommand{\bfn}{{\mathbf{n}}}
\newcommand{\bfo}{{\mathbf{o}}}
\newcommand{\bfp}{{\mathbf{p}}}
\newcommand{\bfq}{{\mathbf{q}}}
\newcommand{\bfr}{{\mathbf{r}}}
\newcommand{\bfs}{{\mathbf{s}}}
\newcommand{\bft}{{\mathbf{t}}}
\newcommand{\bfu}{{\mathbf{u}}}
\newcommand{\bfv}{{\mathbf{v}}}
\newcommand{\bfw}{{\mathbf{w}}}
\newcommand{\bfx}{{\mathbf{x}}}
\newcommand{\bfy}{{\mathbf{y}}}
\newcommand{\bfz}{{\mathbf{z}}}

\newcommand{\tA}{{\widetilde{A}}}
\newcommand{\tB}{{\widetilde{B}}}
\newcommand{\tC}{{\widetilde{C}}}
\newcommand{\tD}{{\widetilde{D}}}
\newcommand{\tE}{{\widetilde{E}}}
\newcommand{\tF}{{\widetilde{F}}}
\newcommand{\tG}{{\widetilde{G}}}
\newcommand{\tH}{{\widetilde{H}}}
\newcommand{\tI}{{\widetilde{I}}}
\newcommand{\tJ}{{\widetilde{J}}}
\newcommand{\tK}{{\widetilde{K}}}
\newcommand{\tL}{{\widetilde{L}}}
\newcommand{\tM}{{\widetilde{M}}}
\newcommand{\tN}{{\widetilde{N}}}
\newcommand{\tO}{{\widetilde{O}}}
\newcommand{\tP}{{\widetilde{P}}}
\newcommand{\tQ}{{\widetilde{Q}}}
\newcommand{\tR}{{\widetilde{R}}}
\newcommand{\tS}{{\widetilde{S}}}
\newcommand{\tT}{{\widetilde{T}}}
\newcommand{\tU}{{\widetilde{U}}}
\newcommand{\tV}{{\widetilde{V}}}
\newcommand{\tW}{{\widetilde{W}}}
\newcommand{\tX}{{\widetilde{X}}}
\newcommand{\tY}{{\widetilde{Y}}}
\newcommand{\tZ}{{\widetilde{Z}}}

	\maketitle

\begin{abstract}%
We study the scaling of classification error rates with respect to the size of the training dataset. In contrast to classical results where rates are minimax optimal for a problem class, this work starts with the empirical observation that, even for a fixed data distribution,  the error scaling can have \emph{diverse} rates across different ranges of sample size. 
To understand when and why the error rate is non-uniform, we theoretically analyze nearest neighbor classifiers.
We show that an error scaling law can have fine-grained rates: in the early phase, the test error depends polynomially on the data dimension and decreases fast; whereas in the later phase, the error depends exponentially on the data dimension and decreases slowly. 
Our analysis highlights the complexity of the data distribution in determining the test error. When the data are distributed benignly, we show that the generalization error of nearest neighbor classifier can depend polynomially, instead of exponentially, on the data dimension. 
\end{abstract}



\section{Introduction}

A scaling law in learning describes the convergence rates of test error with respect to the size of the training data set, the amount of computation, and the size of the model. Remarkably, even for complex models like neural networks, scaling laws often exhibit a fast power-law relationship when applied to real-world tasks such as natural language processing and computer vision~\citep{kaplan2020scaling, zhai2021scaling}. The fast empirical rate raises interesting theoretical questions: Under what conditions, and why, does population risk converge fast with respect to the dataset size without suffering from the curse of data dimensionality? Additionally, is there a limit point at which the scaling law starts to slow down?

This work studies the aforementioned questions by analyzing the population risk as a function of the dataset size $n$ in \emph{binary classification problems}. 
Consider a sample of $n$ independently and identically distributed observations $(X_1,Y_1),\dots,(X_n,Y_n)\in \calX \times \calY$ drawn from an unknown distribution, where $\calY=\{0,1\}$. The goal is to use this data to construct a classifier $\hat f_n:\calX \mapsto \calY$. The \emph{misclassification rate} 
for a classifier $f$ is the population risk using zero-one loss defined as
\begin{equation}
\label{eq:loss}
R(f) \triangleq \Prob[f(X) \ne Y],
\end{equation}
which is also referred to as the \emph{test error}.
The problem has been extensively studied, and the oracle classifier that minimizes the misclassification rate is the well-known Bayes decision rule 
\begin{align}
\label{eq:eta}
f^*(x) = \indc{\eta(x)\ge 1/2}, \quad \text{where }\  \eta(x)\triangleq \Prob[Y=1|X=x].
\end{align}
Here, $\eta$ is the regression function, and the Bayes risk, denoted as $R^*\triangleq R(f^*)$, represents the minimal test error based on infinite training data. As the sample size $n$ increases, the aim is to construct a classifier $\hat f_n$ with a diminishing \emph{excess risk}, defined as
\[
\text{Excess Risk}(f) \triangleq R(f) - R^*.
\]
The excess risk of $\hat f_n$ is a random quantity dependent on the training dataset. In this paper, we focus on the \emph{expected excess risk}. 

Bounding the excess risk of a learned classifier $\hat{f}_n$ is a fundamental topic in statistical learning theory. Most results fall in two categories. The first category is the \emph{parametric rate}. When the complexity of the function class to which the regression function $\eta$ belongs is bounded, empirical process theory can yield rates of the form $\cO (\sqrt{p / n})$ or $\cO (p / n)$, where $p$ measures the complexity of the function class \cite{alon1997scale, bartlett2002rademacher, vapnik2015uniform}. 
The second category is the \emph{non-parametric rate} \cite{Stone1982,tsybakov2009nonparametric,ATW19,SchmidtHieber2020}, which applies to cases where the complexity of the function class---such as nearest neighbors, random forests, or neural networks---can be large or grow with the dataset size. In these problems, the minimax optimal rate for the expected excess risk usually takes the form of $\cO (n^{-\Theta(1/d)})$, where $d$ represents the dimension of the feature space $\calX$, and the hidden constants depend on other parameters such as the margin and smoothness of the regression function~\cite{tsybakov2004optimal,AT07}. 
More precisely, we say an error rate is \emph{parametric} if $ \lim_{n\to \infty} \log(n)/ \log( \Expect R(\hat{f}_n) - R^*)$ is independent of $d$, and an error rate is \emph{nonparametric} if $ \lim_{n\to \infty} \log(n) / \log(\Expect R(\hat{f}_n) - R^*)) $ depends on $d$, e.g., $\Expect R(\hat{f}_n) - R^*  = n^{\Theta(1/d)}$.

\begin{figure}[tb] 
\centering 
\includegraphics[width=0.7\hsize]{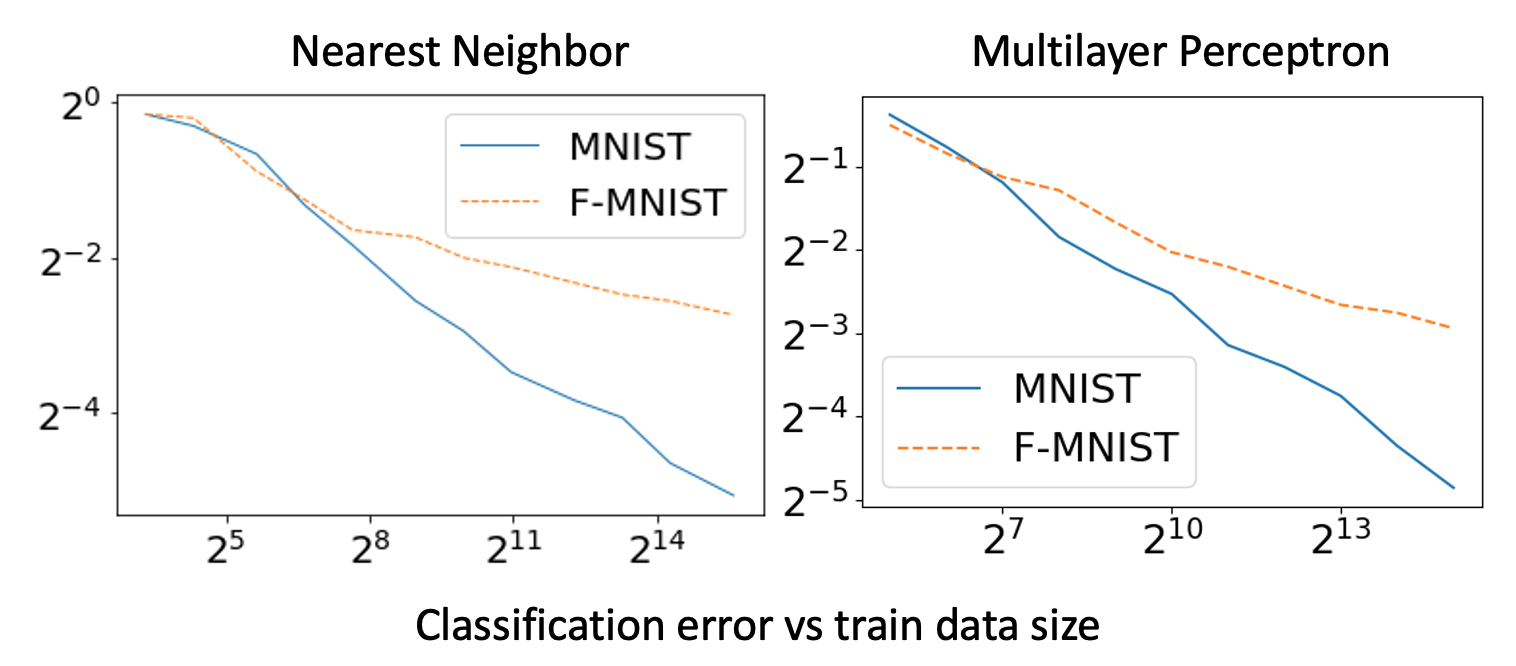} 
\caption{
Test error versus sample sizes on MNIST and F-MNIST datasets.
The error for F-MNIST data initially matches that of the MNIST data but slows down later. This figure shows that the same model can perform differently for very similar classification problems, and hence accurately characterizing the excess risk would require considering the complexity of the data distribution. 
} 
\label{fig:mnist} 
\end{figure}

The aforementioned rates are often minimax optimal, yet they only provide tight upper bounds for the worst-case distributions and do not illuminate on when faster rates can be achieved in practice. 
For example, consider classifiers trained on two image classification datasets, MNIST and F-MNIST, shown in Figure~\ref{fig:mnist}. These experiments reveal two interesting phenomena: 
\begin{itemize}
    \item \emph{Fast rates}. Even though the data dimension is hundreds, nonparametric models can achieve fast rates as in the MNIST experiment \emph{without} suffering from the curse of dimension.
    \item \emph{Multiphase rates}. Unlike most analyses of test error, the error rate can be non-uniform across $n$ as in the F-MNIST experiment. Both fast and slow rates can \emph{coexist} for the same model within different ranges of dataset size $n$. 
\end{itemize}
These phenomena are commonly observed in machine learning practice with state-of-the-art models \cite{kaplan2020scaling,zhai2021scaling}, and can also be similarly reproduced through synthetic experiments under the logistic regression models (see \prettyref{fig:syn}).  
More specifically, we consider the logistic regression model with a normal design: 
\begin{equation}
\label{eq:model-logit}
\Prob[Y=1|X=x] = \frac{1}{1+e^{-x^\top \theta_*/\sigma}},
\quad 
X \sim \calN(0, I_d),
\end{equation}
where $\theta_*\in\reals^d$ represents the optimal decision boundary, and $\sigma=1$ is the noise level. When the data follows standard Gaussian distribution, we see in the \emph{first} row of Figure~\ref{fig:syn} that nearest neighbors learn at polynomial rates. However, when the data becomes uniform over a rectangle (see Section~\ref{sec:multiphase} for more details), the multi-phase rate appears in the \emph{second} row of Figure~\ref{fig:syn}. Motivated by the above observations, our work aims to theoretically analyze when parametric rates can or cannot be achieved by nearest neighbor classifiers. We summarize our results below.



\subsection{Contributions} 
We identify the cause of multiphase rates as the diversity in learnability at different test points $x$. More specifically, we first quantify the pointwise learnability of an instance  $x \in \reals^d$ using $\tau_\rho(x)$ defined in \eqref{eq:def-tau}. With this, we get several conclusions below.

\paragraph{Instance learnability and general error rates.} By characterizing the distribution of the instance learnability through $\psi(\rho, t)$ defined in \eqref{eq:majorizing}, we provide a distribution-dependent error rate (\prettyref{thm:main}) and an improved rate (\prettyref{thm:main-margin}) under the additional margin condition~\eqref{eq:margin}. 


\paragraph{Optimal error rates for the logistic problem with a norm design.} With the above tools, we first analyze the standard logistic regression model (Theorem~\ref{thm:logit}), demonstrating that \kNN\ classifiers can achieve the parametric rate 
\begin{align*}
    \Expect R(\fkNN_{n,k})-R^* = \calO(d/n),
\end{align*}
which matches the minimax lower bound up to constant factors~\cite{hsu2024sample,KvG2024}. Further details are provided in~\prettyref{sec:logit} and~\prettyref{rmk:logit-lower}.

\paragraph{Provable two-phase rates for a modified logistic problem.} In the \emph{second} application, we show that a slight variation of the logistic regression model can produce multiphase rates in~\prettyref{sec:two-phase}. Specifically, when the first two coordinates are distributed as shown in the second row of~\prettyref{fig:syn}, we show in both the experiment (Figure~\ref{fig:syn}) and Theorem~\ref{thm:two-phase-upper}, that the error rate initially follows the parametric rate before eventually slowing to a nonparametric rate. In Theorem~\ref{thm:two-phase-lower}, we prove that this final phase is unavoidably slower than any polynomial rate. 

\paragraph{Connection to previous results.} The instance learnability $\tau_\rho(x)$ studied in this work can be considered an \emph{orthogonal} analysis to the previous smoothness~\eqref{eq:smoothness} or  margin~\eqref{eq:margin}  conditions in analyses of nearest neighbor classifiers. The instance learnability highlights the existence of point-wise error rates in a fixed classification problem, and leads to different convergence rates for different range of sample sizes.

On the one hand, the instance learnability condition (Definition~\ref{def:learnability}) can lead to parametric rates in benign setups, unseen in previous studies (Theorem~\ref{thm:main}, Theorem~\ref{thm:logit}). On the other hand, our condition may further benefit from additional smoothness or margin assumptions (Theorem~\ref{thm:main-margin}).

\subsection{Background on Nearest Neighbor Classifiers}
We first introduce the $k$-nearest neighbor (\kNN) classifier. Given a set of labeled data $\cD_n = \{(X_1,Y_1), \dots,$ $(X_n,Y_n) \} \subseteq \reals^d \times \{0,1\}$, the \kNN\ classifier averages the $k$ closest points to $x$ according to the chosen reference norm $\Norm{X_i-x}$, with distance ties broken uniformly at random. In other word, we reorder the dataset as $(X_{(1)}(x),Y_{(1)}(x)), \dots, (X_{(n)}(x),Y_{(n)}(x))$ such that $\Norm{X_{(i)}(x)-x}$ is increasing. The \kNN\ classifier is then defined as
\begin{equation}
    \label{eq:vanilla-kNN}
    \fkNN_{n,k}(x) \triangleq \indc{\sum_{i=1}^k Y_{(i)}(x)  \ge k/2}. 
\end{equation}
Throughout this paper, we consider the Euclidean norm $\Norm{\cdot}_2$.

Given the \kNN\ classifier defined in~\eqref{eq:vanilla-kNN}, the observed different error rates must stem from differences in data distributions. Significant progress has been made in understanding the consistency and convergence rates of nearest neighbor classifiers.
We present two key conditions below. A common assumption is \emph{H\"older continuity}, which assumes that, for some constant $L>0$ and $0<\beta\le 1$, 
\begin{align}\label{eq:smoothness}
    |\eta(x) - \eta(x')| \le L\|x - x'\|_2^\beta,\quad \forall x,x'.
\end{align}
Under the H\"older continuity, the expected excess risk decays at $\cO(n^{-\frac{\beta}{2\beta + d}})$ for compactly supported features \cite{gyorfi2002distribution}, where $d$ can be improved to the intrinsic dimension of the features~\cite{kpotufe2011,gottlieb2016nearly}, and $L$ can be the average case smoothness~\cite{ashlagi2021functions, hanneke2023near}. In addition to the smoothness condition, recent contributions often incorporate \emph{margin (or low noise) conditions} relevant to plug-in estimators \cite{tsybakov2004optimal,MN06}. One explicit assumption is
\begin{align}\label{eq:margin}
    \Pr\left[0<\abs{\eta(x)-\frac{1}{2}}\le t \right] \le C_0 t^\alpha,\quad \forall~t\in(0,t^*],
\end{align}
for some constants $C_0,\alpha$, and $t^*\le \frac{1}{2}$.
Under this condition, the convergence rate can be improved to $\cO(n^{-\frac{\beta(1+\alpha)}{2\beta + d}})$~\cite{Samworth2012,GKM16}. Additional generalized margin conditions have been explored by works such as~\cite{CD14,xue2018achieving,doring2017rate, chen2018explaining,GW21}.

\begin{figure}[tb] 
\centering 
\includegraphics[width=0.8\hsize]{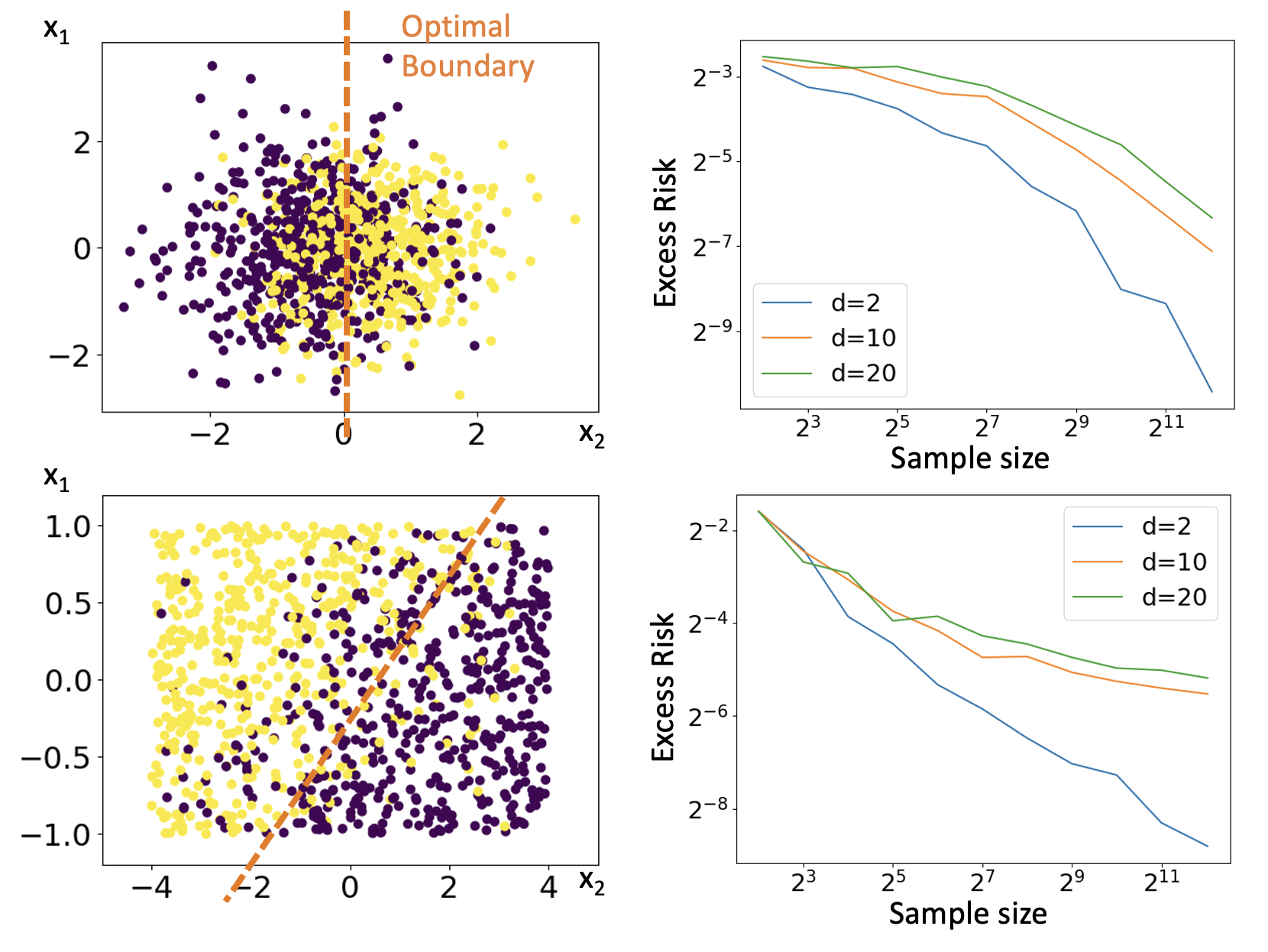} 
\caption{
The \textbf{left} column shows the data distributions in the first two coordinates for the two problems---the logistics regression model and a modified version---introduced in Section~\ref{sec:app}. The remaining coordinates are independent standard Gaussian. The \textbf{right} column shows the excess error of a learned \kNN\ classifier across various problem dimensions and sample sizes. We observe, and will prove in Section~\ref{sec:two-phase}, that the modified logistic regression model leads to multiphase rates in the second row. We show in \prettyref{thm:two-phase-lower} that this modified experiment provably suffers from the curse of dimensionality.
} 
\label{fig:syn} 
\end{figure}

\begin{remark}[Faster rates in known result]
We note that under the additional smoothness and low-noise condition, the convergence rate can be improved to $\cO\left(n^{-\frac{\beta(1+\alpha)}{2\beta + d}}\right)$ or even $\cO\left(\exp{\left(-n c^\frac{\beta(1+\alpha)}{2\beta + d}\right)}\right)$ for $c \in (0,1)$. However, this does not close the gap between parametric and nonparametric rates. Recall the logistic regression model with a normal design in~\eqref{eq:model-logit}.
For this logistic regression model, we observe that $\alpha = 1$ (see~\prettyref{lmm:logit-margin})
and $\beta = 1$. This results in an sample complexity exponentially large  in $d$ for achieving any nontrivial error $\epsilon < 0.5$. However, as shown in the first row of Figure~\ref{fig:syn}, nearest neighbor classifiers can achieve a much faster parametric rate. To theoretically justify the fast rate, we introduce a fine-grained analysis in the next section.  
\end{remark}

\paragraph{Notations}
Let $[n]\triangleq \{1,\dots,n\}$.
Let $B_r(x)\triangleq \{y: \Norm{y-x}_2<r \}$ denote the Euclidean open ball of radius $r$ centered at $x$.
For $x\in\reals^d$, let $(X_{(1)}(x),Y_{(1)}(x)), \dots, (X_{(n)}(x),Y_{(n)}(x))$ denote a reordering of $(X_1,Y_1), \dots,(X_n,Y_n)$ such that $\Norm{X_{(i)}(x)-x}_2$ is increasing, and let $R_{(i)}(x)\triangleq \Norm{X_{(i)}(x)-x}_2$.
For $x,y\in\reals$, let $x\wedge y \triangleq \min\{x,y\}$ and $x\vee y \triangleq \max\{x,y\}$.
For two positive sequences $\{a_n\}$ and $\{b_n\}$, we write $a_n=\cO(b_n)$ or $a_n\lesssim b_n$ if there exists a constant $C$ such that $a_n\le C b_n$, and we write 
$a_n\asymp b_n$ if $a_n\lesssim b_n$ and $b_n\lesssim a_n$.

\section{Fine-grained Error Rates}
\label{sec:fine-grained}

In this section, we formally introduce our analyses which permits fine-grained error rates that are adaptive to the structure of the data distribution. We begin by examining the \kNN\ classifiers from a Bayesian statistics perspective to describe the data distribution. 

The predictor that minimizes the misclassification rate is known to be the \emph{maximum a posteriori} (MAP) estimator:
\begin{equation}
\label{eq:MAP}
f^*(x)=\argmax_{y\in\calY} \Prob[Y=y|X=x].
\end{equation}
However, the MAP estimate is generally not feasible because the underlying data distribution is unknown, and estimating the conditional distribution $P_{Y|X=x}$ is often challenging. 
Instead of the conditional distribution $P_{Y|X=x}$,
consider the \emph{smoothed MAP} estimation with a bandwidth $r$:
\begin{equation}
\label{eq:MAP-smooth}
f_r^*(x)=\argmax_{y\in\calY} \Prob[Y=y|X\in B_r(x)].
\end{equation}


Using an empirical estimate of the conditional distribution, the smoothed MAP with a bandwidth $r$ corresponds to the majority vote among observations in $\cD_n$ for which $X_i\in B_r(x)$. The bandwidth $r$ must be chosen to ensure that $P_X(B_r(x))$ is sufficiently large. A data-driven choice is the distance to the $(k+1)\Th$ nearest neighbor of $x$, which leads to the \kNN\ classification rule.

\subsection{Pointwise Error Probabilities}
\label{sec:pointwise}
\begin{figure*}[t] 
\centering 
\includegraphics[width=\hsize]{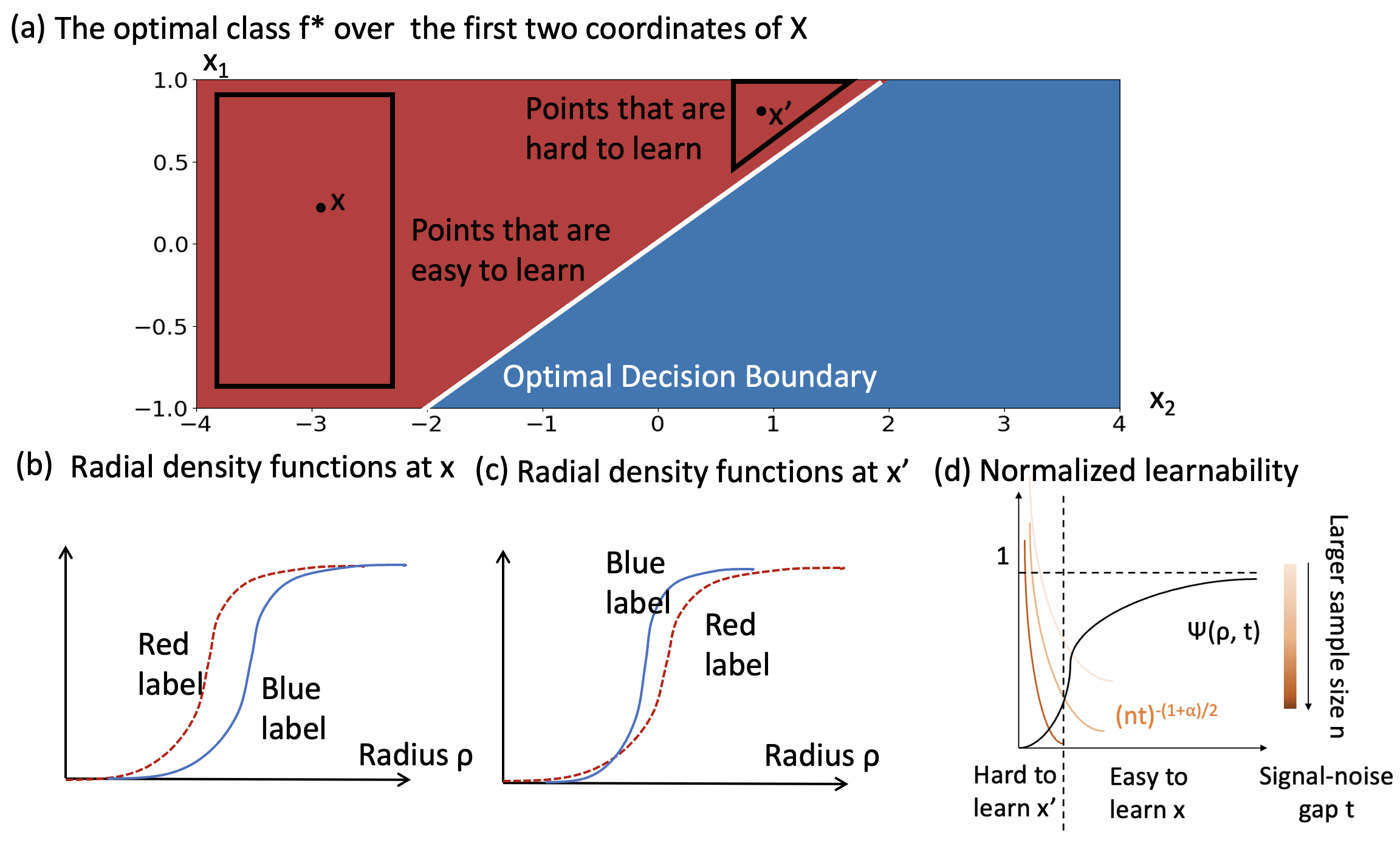} 
\caption{We illustrate the modified logistic model in Figure~\ref{fig:syn}. (a) The data distributes uniformly. The blue and the red regions correspond to the optimal predictor $f^*(x) = 0, 1$ respectively. The exact distribution can be found in Section~\ref{sec:multiphase}. 
(b)(c) We plot the radial density function at point $x$ and $x'$ using the radius $r$ computed in the first two coordinates. The plot in (b) shows that the red curve always dominates the blue curve. The difference between (b) and (c) illustrate why two-phase rates can happen in the same model. (d) An illustration for the right-hand sides of Theorems~\ref{thm:main} and~\ref{thm:main-margin}. The function $\psi(\rho, t)$ defined in \eqref{eq:majorizing} is monotonically increasing in $t$. The intersection point determines the order of the expected excess risk, illustrating how the error decreases as the sample size $n$ increases.
} 
\label{fig:tau} 
\end{figure*}

We start our analysis by examining the second experiments in Figure~\ref{fig:syn}, where the first two coordinates are uniform over a rectangle, and all other coordinates are standard Gaussian. We illustrate the first two coordinates in Figure~\ref{fig:tau} a . The label distribution is logistic and can be found later in Equation~\eqref{eq:logit-modified}. 
We observe that multiple error rates can coexist in a error scaling law because the learnability can vary between instances (i.e. test points) $x$, reflecting the practical observation that some instances are easier (Figure~\ref{fig:tau} b), while others are more challenging (Figure~\ref{fig:tau} c). 

By leveraging the connection between \kNN\ classifiers and the smoothed MAP estimate, we derive pointwise error probabilities $\Prob[\fkNN_{n,k}(x)\ne f^*(x)]$ for \kNN's prediction on $x$ with reference to $f^*(x)$. 
For a fixed radius $r$, the smoothed MAP~\eqref{eq:MAP-smooth} defines the relative signal strength from the reference label $f^*(x)$ as the ratio 
\begin{equation}
\label{eq:signal-relative}
S_r(x)
\triangleq 
\frac{\Prob[Y=j^*|X\in B_r(x)]}{\Prob[Y=j|X\in B_r(x)]},
\quad \text{ if } f^*(x)=j^*, j\ne j^*.
\end{equation}
For a data-driven radius $r$, its typical values are relevant. In particular, if $r=R_{(k+1)}(x)$ is the distance to the $(k+1)\Th$ nearest neighbor to $x$, then, by applying concentration inequalities, the total mass of the ball $B_r(x)$ is of order $\Theta(k/n)$ with high probability. The following instance learnability $\tau_\rho(x)$ shifts and rescales the relative signal strength over typical radii. 

\begin{definition} 
    \label{def:learnability}
    For $\rho\in(0,1)$ and $x\in \reals^d$, let\footnote{
    The factor 2 is for notion convenience only and can be replaced by any constant. 
    } the instance learnability be defined as follows,
    \begin{equation}
    \label{eq:def-tau}
    \tau_\rho(x) \triangleq 
    \inf\sth{ \sqrt{\rho}\pth{S_r(x)-1}:  \frac{\rho}{2} \le P_X(B_r(x)) \le 2\rho  }.     
\end{equation}
\end{definition}

If $S_r(x)>1$ over typical radii, and thus $\tau_\rho(x)>0$, then $\cD_n$ satisfying $X_i\in B_r(x)$ is dominated by instances with the label $f^*(x)$. Consequently, the \kNN\ classifier with $k=\Floor{n\rho}$ aligns with the MAP estimate with high probability, as shown in~\prettyref{prop:pointwise}. 
The proof applies the conditional distribution of the order statistics from~\cite{KR92} and the fine-grained concentration inequality in~\cite{Hoeffding94}. The proof details are provided in Appendix~\ref{sec:pf-general}.
\begin{prop}
    \label{prop:pointwise}
    If $\tau_\rho(x)\ge 0$ and $k=\Floor{n\rho}\ge 5$, then
    \begin{equation}
    \label{eq:error-pointwise}
    \Prob[\fkNN_{n,k}(x)\ne f^*(x)]
    \le 3\exp\pth{- \frac{1}{25}n\pth{\tau_\rho^2(x)\wedge \rho} }.
    \end{equation}
\end{prop}
This proposition explains why the error rate at some instance $x$ can decay polynomially. In Figure~\ref{fig:tau}(a)(b), we see that at point $x$,  $S_r(x)\ge1$ for any $r$, and hence the effective radius $\rho$ doesn't have to be very small for the algorithm to learn.

Conversely, if $S_r(x)<1$ over typical radii (as in Figure~\ref{fig:tau} c), then $\cD_n$ satisfying $X_i\in B_r(x)$ is dominated by instances with a label different from $f^*(x)$. In this case, since $\liminf_{r\to 0} S_r(x)\ge 1$ for smooth $\eta$, a larger sample size $n$ is required to ensure that $\rho\ge 1/n$ is sufficiently small, thereby guaranteeing $\tau_\rho(x)\ge 0$. Consequently, due to such instances, an eventual slower convergence rate could be unavoidable. We will provide further details on the impossibility in \prettyref{sec:two-phase}.


\begin{remark}[General metric spaces and multiple classes]
This paper focuses on the binary classification in Euclidean space, but the results can readily extend to general spaces, provided an appropriate metric is specified for nearest neighbor methods. For multiclass classification, the smoothed MAP~\eqref{eq:MAP-smooth} defines the relative signal strength as the minimum ratio over all pairwise comparisons:
\[
S_r(x)=
\min_{j\ne j^*}\frac{\Prob[Y=j^*|X\in B_r(x)]}{\Prob[Y=j|X\in B_r(x)]},
\quad \text{ if } f^*(x)=j^*,
\]
in place of~\eqref{eq:signal-relative}. The analysis can then be generalized accordingly. 
\end{remark}

\subsection{Excess Risk Bounds via Normalized Learnability}
\label{sec:general}
The diverse error rates of $\fkNN_{n,k}$ arise from the distribution of instance learnability, applying~\eqref{eq:error-pointwise} to integrate pointwise error probabilities over the feature distribution. Furthermore, the expected excess risk can be represented as the \emph{weighted} average of pointwise error probabilities: 
\begin{equation}
\label{eq:excess-integral}
\Expect R(\fkNN_{n,k}) - R^*
=\int |2\eta(x)-1| \Prob[\fkNN_{n,k}(x)\ne f^*(x)] P_X(\diff x).
\end{equation}
From this representation, to achieve a desired error, the learnability $\tau_\rho(x)$ can be smaller if the regression function $\eta(x)$ is close to $1/2$. This motivates the following increasing subsets of features based on \emph{normalized} learnability:
\begin{equation}
    \label{eq:E-rho-t}
    E_\rho(t) \triangleq \sth{x: \frac{\tau_\rho(x)}{|\eta(x)-\frac{1}{2}|} \le t}.
\end{equation}
If $x\in E_\rho(t)$ for a large $t$, its contribution to the weighted average~\eqref{eq:excess-integral} will be small.

Now we introduce our characterization of data distribution complexity and the associated risk bounds. As seen from~\eqref{eq:excess-integral}, the distance of the regression function from $1/2$ and the probability mass of $E_\rho(t)$ together determine the expected excess risk. The complexity is captured by a majorizing function $\psi(\rho,t)$, which bounds the product of $\sup_{x\in E_\rho(t)} |2\eta(x)-1|$ and $P_X(E_\rho(t))$:
\begin{equation}
    \label{eq:majorizing}
    \sup_{x\in E_\rho(t)} |2\eta(x)-1| \cdot P_X(E_\rho(t)) \le \psi(\rho,t),
\end{equation}
where the supremum over an empty set is defined as zero.

\begin{theorem}
    \label{thm:main}
    If $\psi$ is a majorizing function satisfying~\eqref{eq:majorizing} and $k=\Floor{n\rho}\ge 5$, then 
    \begin{equation}
        \label{eq:thm-main}
        \Expect R(\fkNN_{n,k}) - R^*
        \le \inf_{t>0} \sth{\psi(\rho,t) + \frac{30}{\sqrt{2e n t^2}}}  + 3 e^{-\frac{n\rho}{25}}. 
    \end{equation}
\end{theorem}

With the additional margin condition~\eqref{eq:margin}, the second term in~\eqref{eq:thm-main} can be further improved.

\begin{theorem}
    \label{thm:main-margin}
    If $\psi$ is a majorizing function satisfying~\eqref{eq:majorizing}, the margin condition~\eqref{eq:margin} holds, and $k=\Floor{n\rho}\ge 5$, then
    \begin{equation}
        \label{eq:thm-main-margin}
        \Expect R(\fkNN_{n,k}) - R^*
        \le \inf_{t>0}\sth{ \psi(\rho,t) 
        + C_\alpha \pth{\frac{5}{\sqrt{n t^2}}}^{1+\alpha} }
        + 3 e^{-\frac{n\rho}{25}},
    \end{equation}
    where $C_\alpha$ depends only on $\alpha$.
\end{theorem}



Theorems~\ref{thm:main} and~\ref{thm:main-margin} upper bound the expected excess risk as a sum of three terms. The first term captures the risk due to instances with normalized learnability below $t$; the second term bounds the test error for instances with normalized learnability above $t$; and the third term represents large deviations for the tail event when the data-driven radius $R_{(k+1)}(x)$ falls outside its typical range.

The proofs essentially follow the above interpretation. 
The proof of Theorem~\ref{thm:main} partitions the input domain based on the normalized learnability $t$. For values exceeding $t$, the pointwise error probabilities are upper bounded by~\prettyref{prop:pointwise}.
Under the additional margin conditions, the proof of Theorem~\ref{thm:main-margin} further applies the peeling device (see, e.g.,~\cite{Geer_empirical_2009}) to partition the input domain based on an exponential grid on $|\eta(x)-\frac{1}{2}|$.  
Detailed proofs are provided in Appendix~\ref{sec:pf-general}.

While the proof ideas of these theorems are straightforward, the characterization $\psi(\rho,t)$ provides a novel perspective on the complexity of the data distribution. 
To derive explicit error rates, it suffices to find $\psi(\rho,t)$ at a specific level of $t$. Without loss of generality, we assume that $\psi(\rho,t)\in[0,1]$ and increases with $t$, as both factors on the left-hand sides of~\eqref{eq:majorizing} do. By monotonicity, as $n$ grows, the optimal choice of $t$ decreases, consequently reducing $\psi(\rho,t)$, as illustrated in Figure~\ref{fig:tau}. Therefore, Theorems~\ref{thm:main} and~\ref{thm:main-margin} deliver our fine-grained results on the learning curve.

Due to space constraints, we present further results on the general non-parametric, parametric, and multiphase rates in Appendix~\ref{sec:rate-general}, applicable to diverse $\psi$ functions.
Our results not only recover existing slow rates through \emph{local smoothness} analysis but also yield novel fast and multiphase rates via \emph{global approximation} of the input distribution. 
In the next section, we demonstrate nontrivial global analyses of the $\psi$ function using Gaussian approximation in the context of logistic regression---one of the fundamental models for classification tasks.




\section{Fast and Multiphase Rates}
\label{sec:app}

We apply the theorems above to two examples. The first example is a standard logistics regression model. Although nearest neighbor classifiers are agnostic to the problem structure, they can achieve statistically optimal error rates $\calO(d/n)$. The second example shows that a multiphase rate can be observed and provably happens for a slightly modified logistic regression model.

\subsection{Fast Rates for Standard Logistic Regression Model}
\label{sec:logit}

In this subsection, we study the standard logistic regression model~\eqref{eq:model-logit} for binary classification. The maximum likelihood estimation (MLE) finds the coefficient $\hat \theta$ that maximizes the log-likelihood $\sum_{i=1}^n - \log(1+ \exp(-(2Y_i-1)X_i^\top \theta))$, with the predictor given by $\hat f(x)=x^\top \hat \theta$. However, the MLE does not exist when the data are linearly separable~\cite{AA1984,CS2020}, necessitating additional regularization~\cite{KvG2024}.

In contrast, the nearest neighbor method is agnostic to the underlying structure, and its predictor is always well-defined. Although local methods can suffer from the curse of dimensionality in the worst case, they may still be \emph{adaptive} to the underlying structure, as shown in Figures~\ref{fig:mnist} and \ref{fig:syn}. \cite{kpotufe2011} demonstrates that $\fkNN_{k,n}$ is adaptive to the intrinsic dimension of the features $X$; however, here $X$ is fully supported, with only $P_{Y|X}$ exhibiting a simple structure. 

The following theorem provides a resolution for the fast rate that is consistent with empirical observations. In the statement, let $\calL_d$ denote the family of all distributions for the logistic regression model~\eqref{eq:model-logit} with $\theta_*\in S^{d-1}$ and $\sigma=1$. 

\begin{theorem}
    \label{thm:logit}
    There exist constants $c_0, c,C$ such that, if $k=\Floor{c_0 n}$ and $d \ge C \log n$, then
    \[
    \sup_{P\in \calL_d} \Expect R(\fkNN_{n,k}) - R^*
    \le \frac{c d}{n}.
    \]
    Furthermore, there exists a constant $c'$ such that
    \[
    \inf_{\hat f}\sup_{P\in \calL_d} R(\hat f) - R^* \ge \frac{c'd}{n}.
    \]
\end{theorem}


\paragraph{Proof Sketch.} The main challenge in establishing the fast rate is to show that the logistic regression model satisfies~\eqref{eq:majorizing} with 
\begin{equation}
\label{eq:psi-logit-1}
\psi\pth{c_0,\sqrt{c/d}} \le \cO(d/n).
\end{equation}
Since the margin condition~\eqref{eq:margin} holds with $\alpha=1$ (see \prettyref{lmm:logit-margin}), applying~\prettyref{thm:main-margin} yields the desired result. 
The key structural property of the data distribution is as follows (see \prettyref{prop:logit-dominance}). 
Then~\eqref{eq:psi-logit-1} follows from the Bernstein inequality.
\begin{lemma}
    \label{lmm:logit-dominance-main}
    If $\Norm{x}_2 \lesssim\sqrt{d}$ and $\rho\le \frac{1}{12}$ is a constant, 
    then
    \begin{equation}
    \label{eq:tau-rho-goal}
    \tau_{\rho}(x) \gtrsim (|x^\top \theta_*| \wedge 1)/\sqrt{d}.
    \end{equation}
\end{lemma}

To prove~\prettyref{lmm:logit-dominance-main}, one technical innovation is the Gaussian approximation for the conditional distributions (see Lemma~\ref{lmm:gaussian-approx}).
Specifically, let $\mu_{i,r}(x)\triangleq \Prob[X\in B_r(x) \mid Y=i]$.
Consider $f^*(x)=1$. The goal is to lower bound
\begin{equation}
\label{eq:Srx}
S_r(x)-1= \frac{\mu_{1,r}(x)-\mu_{0,r}(x)}{\mu_{0,r}(x)}
\end{equation}
uniformly over the typical radii.
Applying the Berry-Esseen bound, the denominator can be approximated by the Gaussian CDF at some standardized point $z$ with $\Phi(z)\approx \rho$ due to the constraint in~\eqref{eq:def-tau}.
However, this approximation is too crude for the numerator which involves a difference of two terms.
Instead, we approximate the \emph{probability density function} (PDF) using the Gaussian PDF and derive a lower bound on the integral. 
By the local limit theorem, the density function can be approximated by $\frac{1}{v(x)}\phi(z)$, where $v^2(x)$ denotes the corresponding variance of order $\cO(d)$. 
Then, \eqref{eq:Srx} is approximately
\[
S_r(x)-1
\approx \frac{\frac{1}{v(x)}\phi(z)\cdot I(x)}{\Phi(z)}
\gtrsim \frac{I(x)}{\sqrt{d} \cdot |z|},
\]
where we applied Mill's ratio and $I(x)$ is a lower bound on the integrated signal (see \eqref{eq:lb-numer-1} and \eqref{eq:lb-numer-large-1}).
By analyzing the stochastic dominance property and applying quantile coupling between the distributions on the signal components $\theta_* \theta_*^\top (X-x)$, we establish a lower bound of order $\Omega(x^\top \theta_*\wedge 1)$ on the signal strength (see \prettyref{lmm:quant-dominance}), thereby obtaining a lower bound for~\eqref{eq:Srx}.
Our analysis reveals that the $\sqrt{d}$ factor in~\eqref{eq:tau-rho-goal} arises from the standard deviation $v(x)$ rather than a direct count of the number of variables. 
The complete proof is detailed in~\prettyref{sec:dominance}. 

\begin{remark}[Extensions: model, neighbor order, noise level]
We focus on logistic regression with unit signal-to-noise ratio and $k\asymp n$ for clarity. The results extend to broader settings:
\begin{itemize}
    \item Similar analyses apply to generalized linear models with $\Prob[Y=1|X=x]=F(x^\top \theta_*/\sigma)$, where $F: \reals\mapsto [0,1]$ is a smooth link function, such as the cumulative distribution function (CDF) of the standard normal distribution in probit models. 
    \item The results extend to general $k$ provided the Gaussian approximation remain valid. Both analytical and numerical results reveal a convergence rate of $\cO(d/k)$. The sample complexity remains polynomial in $d$ when $k= \text{poly}(n)$.
    \item The expected excess risk depends on the noise level through the factor $|2\eta(x)-1|$ in~\eqref{eq:excess-integral}. In the noiseless case with $\eta(x)\in\{0,1\}$, the rate is $\cO(\sqrt{d/n})$ as shown in the next remark. 
\end{itemize}

\end{remark}

\begin{remark}[Minimax optimality]\label{rmk:logit-lower}
\prettyref{thm:logit} establishes minimax optimality in terms of the expected excess risk, represented as the weighted average of error probabilities in~\eqref{eq:excess-integral}. Our pointwise analysis yields valuable insights that applies directly to other key error metrics. One notable application is the \emph{unweighted} average error discussed in~\cite{tsybakov2004optimal}:
\[
d_\triangle(\hat f, f^*)\triangleq 
\int \indc{\hat f(x)\ne f^*(x)} P_X(\diff x).
\]
When $\hat f=\indc{x^\top\hat\theta>0}$ is a linear classifier with a separating hyperplane $\hat \theta$, under Gaussian design, the expected error $\Expect d_\triangle(\hat f, f^*)$ is of the same order as the estimation error $\Expect\Norm{\hat \theta-\theta_*}_2$. Our pointwise analysis shows that, by choosing $k=\Floor{c_0 n}$, the expected error is $\Expect d_\triangle(\fkNN_{n,k}, f^*)\lesssim \sqrt{d/n}$, matching the minimax lower bound for the estimation error~\cite{KvG2024,hsu2024sample} within constant factors. 
\end{remark}

\begin{remark}[On $d\ge C\log n$]
\label{rmk:d-logn}
\prettyref{thm:logit} imposes the additional condition that $d\ge C\log n$, which is relatively mild for high-dimensional problems. The condition is essential in our analysis to ensure the validity of the Gaussian approximation with $(d-1)$ degrees of freedom. Specifically, the lower bound~\eqref{eq:lb-numer-large-1} in the large signal regime becomes trivial when $d$ is small, as the noise variance becomes dominated by the signal term $(x^\top \theta_*)^2$. We conjecture that this condition is an artifact of our analysis and that the result generally holds for all $d\in\naturals$. A potential approach to addressing this would be to rigorously prove that the pointwise error probability $\Prob[\fkNN_{n,k}(x)\ne f^*(x)]$ decreases monotonically as the signal strength $(x^\top \theta_*)^2$ increases.
\end{remark}

\subsection{Multiphase Rates for Modified Logistic Regression Model}
\label{sec:two-phase}

We analyze the example in the second row of Figure~\ref{fig:syn}. In particular, we show why such a simple example exhibits a two-phase rate: the error vanishes fast at first but significantly slows down later.

\paragraph{Model} 
The model is similar to the logistic model described in~\prettyref{sec:logit}, except that the first two coordinates $X^{(1)}, X^{(2)}$ is uniformly sampled on a bounded rectangle. The other $d-2$ coordinates still follow the standard Gaussian distribution $X^{(i)} \sim \calN(0, 1), i=3,\dots, d$.  
Consider the second row of Figure~\ref{fig:syn}, where the rectangle is $[-1, 1] \times [-4, 4]$, and the label $Y$ is determined by the first two coordinates: 
\begin{equation}
\label{eq:logit-modified}
\Prob[Y=1 | X =x] = \frac{1}{1+\exp\pth{x^{(2)}-2 x^{(1)}}}.
\end{equation}
The optimal predictor $f^*(x)$ is visualized in Figure~\ref{fig:tau}. 

Here, the constants $1$ and $4$ are arbitrarily selected for simplicity. As will be evident later from the analysis, the two-phase rates exist as long as the shape is a rectangle such that the majorizing function $\psi$ admit a multiphase bounds as in \prettyref{sec:multiphase}. The turning point is earlier when the length is much longer than the width.
Although this model is very similar to the one in Section~\ref{sec:logit}, it leads to a two-phase convergence, as stated below.


\begin{theorem}[Multi-phase rates]\label{thm:two-phase-upper}
    For $k= n^{3/4}, d > 10$ and then the excess risk for the model introduced above can be bounded as follows,
    \[
     \Expect R(\fkNN_{n,k}) - R^*
    \le 
    \begin{cases}
      0.45 + \tfrac{Cd}{n^{3/4}}, & \text{if } n \lesssim d^2\\
      C'n^{-1/4d} &  \text{otherwise}.
    \end{cases}       
    \]
\end{theorem}


We see from the above theorem that when the excess error is above a dimension-independent threshold $ 0.45$, the scaling law vanishes polynomially fast when $d$ is large; when the error is below the threshold, the scaling law slows down to exponentially slow rate. The constant $C = 0.45$ is not optimized. The choice that $k = n^{3/4}$ can be generalized to any choice $n^{\gamma}, \gamma \in (1/2, 1)$, and the analysis would be similar.

\paragraph{Proof Sketch.} The proof of Theorem~\ref{thm:two-phase-upper} relies on analyzing the function $\psi(\rho, t)$ in Theorem~\ref{thm:main-margin}. We show that for any point $x$ inside the rectangle
\begin{align*}
    S = \{x \in \calX_1 : \ x^{(1)} \in [-0.8, 0.8], x^{(2)} \in [-2.2, -3.8] \}.
\end{align*}
from Figure.~\ref{fig:tau}(a), the ratio $\tau_\rho(x)$ has a positive margin by analyzing the radial density function in Figure.~\ref{fig:tau}(b). Hence, we get fast rates in the first phase. For the second phase, we show that globally for any $x$, $\psi(\rho, \sqrt{\rho})$ scales down as $\rho^{1/d}$. Together,  for $\rho = n^{-1/4}$, we have that $ \psi(\rho, t) \le 0.45 \wedge \rho^{1/d}.$
Then the result follows from Theorem~\ref{thm:main-margin}. More details can be found in Section~\ref{sec:pf-logit-modified}.

The above result shows that nearest neighbor classifiers can provably converge fast in the early phase, while maintaining a slow nonparametric in the second phase, as observed in Figure~\ref{fig:syn}. 
Next, we further show that the second slow-phase is provably unavoidable.

\begin{theorem}[A lower bound]\label{thm:two-phase-lower}
    There exist absolute constants $c_1,c_2>0$ such that,
    if $n\le e^{c_1 d/\log d}$, then, for any choice of $k\in [n]$,
        \[
        \Expect R(\fkNN_{k,n}) - R^*
        \ge c_2.
        \]
\end{theorem}

\paragraph{Proof Sketch:} Our analysis upper bounds the signal-to-noise ratio $\tau_\rho(x)$ for $x$ in the triangle 
\begin{align*}
    S' = \{x \in \calX_1 :  2x^{(1)} > x^{(2)} + 0.1 > 1.1 \},
\end{align*}
illustrated in Figure~\ref{fig:tau}. The key challenge in this proof is that we need to upper-bound exponentially small Gaussian-approximation probability error, so that we may conclude exponentially many sample is required (Lemma~\ref{lemma:lower-two-phase}). Such delicate error control cannot be concluded from conventional Berry-Esseen bounds, where the error term is polynomially small. Instead, we prove a Mills-ratio like condition and use multiplicative error (rather than the additive error in Berry-Esseen bounds) to bound the tail error (Lemma~\ref{lmm:noncentral}).
The complete proof can be found in Section~\ref{sec:pf-logit-modified}. 

\section{Conclusions and Discussions}

This work proposed a fine-grained framework for analyzing test error of nearest neighbor classifier for a given data distribution. With this framework, we can obtain parametric rates when nearest neighbor classifiers are applied to benign problems, and multi-phase error rates, as observed widely in real-world experiments. Our work suggest that test error rates are not uniform across different range of sample sizes.

Multiple extensions and generalizations can be valuable for future studies. One direction is to design better nearest neighbor classifiers that can exploit our analysis. A potential method could be reweighting the samples and another fix might be adaptively choosing the number of neighbors according to the collected information at the test point $x$.  Aside from studying nearest neighbor classifiers, another important and challenging direction is studying the scaling laws for other nonlinear models such as decision trees and neural networks.



\bibliographystyle{alpha}
\bibliography{main}
\newpage

\appendix

\section{Additional Related Works}
The expected excess risk has been widely studied in statistical learning theory, particularly in the context of generalization error. Below, we provide a concise overview of previous results on generalization error and additional related literature on nearest neighbor methods. 

\paragraph{Generalization error.}
Classical approaches often rely on uniform concentration inequalities to provide distribution-free guarantees~\cite{boucheron2005theory, harvey2017nearly, golowich2020size}. The line of research yields the conventional trade-off between approximation error and estimation error, governed by the complexity of the function classes~\cite{VC74,BEHW89,Haussler92}. However, these approaches fail to explain the success of expressive models, notably neural networks~\cite{zhang2021understanding}, overparameterized linear models~\cite{nagarajan2019uniform}, and nearest neighbor classifiers.

Stability-based analyses attempt to address the limitations of uniform concentration approaches by leveraging the following observation:
if the dependency between the input data and the model parameters produced by a learning algorithm is weak, then the generalization gap---the difference between test error and training error---tends to be small. This insight has been instantiated through stability measures~\cite{bousquet2002stability, hardt2016train, bassily2020stability}, information-theoretical measures~\cite{russo2016controlling,xu2017information,haghifam2021towards}, and PAC-Bayesian frameworks~\cite{mcallester1999pac, dziugaite2017computing, neyshabur2017pac}. However, most results do not apply to \kNN\ classifiers that lack explicit parameters. 

Another related approach suggests that if the predicted values are robust to small perturbations in model parameters, the generalization gap will be small. This robustness can be reflected in the flatness of the prediction function~\cite{neyshabur2017exploring}, the margin of the prediction function~\cite{shalev2014understanding, bartlett2017spectrally}, and the norm of the parameters~\cite{gunasekar2018characterizing,wei2019regularization}. However, most results either require positive homogeneous classifiers, such as linear models and ReLU-based neural networks, or impose assumptions that are difficult to verify. 

In summary, general theories on generalization error cannot explain the fast rates observed in experiments with \kNN\ classifiers. Next, we turn to the nearest neighbor literature for more relevant results. 



\paragraph{Nearest neighbors.}
The theoretical study of nearest neighbor methods dates back to \cite{cover1967nearest, Stone1977, fix1989discriminatory}, which established conditions for achieving asymptotic consistency. Later research explored various aspects of nearest neighbor methods, such as strong universal consistency~\cite{DGKL94}, application to general metric spaces~\cite{CG06}, and choice of neighbor order~\cite{HPS08}.

The rate of convergence has been extensively investigated under various smoothness conditions including H\"older continuity~\eqref{eq:smoothness}, yielding asymptotic expansions and the minimax optimal rates \cite{gyorfi1981rate, kulkarni1995rates, SV1998, gyorfi2002distribution}. These results underscore the \emph{curse of dimensionality}, wherein the sample size must grow \emph{exponentially} with the problem dimensionality to achieve consistent estimates of the regression function and diminishing excess risk.

Identifying conditions for faster convergence rates has been the focus of more recent literature. When features exhibit low-dimensional structures, the dimension $d$ can be reduced to the intrinsic dimension of the feature space~\cite{kpotufe2011,gottlieb2016nearly}. Other studies have exploited various margin (or low noise) conditions, including~\eqref{eq:margin}, achieving minimax optimal rates over the corresponding distribution space~\cite{CD14,GKM16,doring2017rate,KpotufeMartinet2021}. We also refer the readers to~\cite{GW21} for an informative summary of these conditions. 

Many recent works also investigate variants of the nearest neighbor methods to achieve stronger guarantees~\cite{Samworth2012, rimanic2020convergence, CBS20, GW21}. However, no known rates have been established with polynomial dependence on the problem dimension. 


\section{General Error Rates}
\label{sec:rate-general}

We present implications of the fine-grained error rates in~\prettyref{sec:fine-grained}, and the corresponding minimax lower bounds. 
The proofs are provided in~\prettyref{sec:pf-rates}
\subsection{Nonparametric Rates}
\label{sec:nonparametric}
The general bounds on expected excess risk encompass many existing convergence rates under smoothness and margin conditions. Below we present examples that derive corresponding majorizing functions $\psi$ that \emph{simultaneously} hold for a family of distributions, which we refer to as \emph{envelope majorizing functions}.

\begin{itemize}
\item \emph{H\"older continuity and strong density condition} \cite{AT07,KK2007,GKM16}.
If the distribution satisfies $\beta$-H\"older continuity~\eqref{eq:smoothness}, the $\alpha$-margin condition~\prettyref{eq:margin}, and the strong density condition $P_X(B_r(x))\ge \kappa r^d$, then it satisfies~\eqref{eq:majorizing} with
\begin{equation}
\label{eq:psi-Holder}
\psi\pth{\rho,\sqrt{\rho}} = C \rho^{\frac{\beta(1+\alpha)}{d}},
\end{equation}
where $C=C_0 \pth{2L \pth{2/\kappa}^{\beta/d}}^{1+\alpha}$.
\item \emph{Generalized Lipschitz condition} \cite{GW21}. 
If the $\alpha$-margin condition~\prettyref{eq:margin} holds and there exists an increasing function $h$ with $\lim_{s\downarrow 0}h(s)=0$ such that $|\eta(x)-\eta(y)|\le h(P_X(B_{\Norm{x-y}_2}(x))$, then the distribution satisfies~\eqref{eq:majorizing} with $\psi(\rho,\sqrt{\rho})=C_0 \pth{4 h(2\rho)}^{1+\alpha}$. A special case, $h(s)=C^* s^{\beta/d}$, is known as the \emph{modified Lipschitz condition}~\cite{doring2017rate}, yielding
\begin{equation}
\label{eq:psi-general-Lip}
\psi\pth{\rho,\sqrt{\rho}} = C' \rho^{\frac{\beta(1+\alpha)}{d}},
\end{equation}
where $C'=C_0\pth{4 C^*\cdot 2^{\beta/d} }^{1+\alpha}$.
\end{itemize}
 
Formal proofs for the validity of these majorizing functions are provided in~\prettyref{sec:pf-rates}. In these proofs, we identify the set of instances $E_\rho(\sqrt{\rho})$ below normalized learnability $\sqrt{\rho}$ under these local smoothness conditions. Our characterization aligns with the set of instances below a certain margin threshold, as in~\eqref{eq:margin}. Thus, this margin threshold can be interpreted as the local normalized learnability as $\rho\downarrow 0$.

To present the general nonparametric rates, consider a majorizing function of the following form:
\begin{equation}
    \label{eq:psi-nonparametric}
    \psi(\rho,t)
    = 
    \begin{cases}
        C_1 \rho^{\delta},& t\le \sqrt{\rho},\\
        1, & t>\sqrt{\rho}.
    \end{cases}
\end{equation}
The next result is derived by applying~\prettyref{thm:main-margin} with $t=\sqrt{\rho}$ and $\rho=n^{-\frac{1+\alpha}{1+\alpha+2\delta}}$.

\begin{corollary}
    \label{cor:nonparametric}
    If $\psi$ specified in~\eqref{eq:psi-nonparametric} is a majorizing function satisfying~\eqref{eq:majorizing}, the margin condition~\eqref{eq:margin} holds, and $k=\Floor{n^{\frac{2\delta}{1+\alpha+2\delta}}}\ge 5$, then    
    \begin{equation}
        \label{eq:ub-nonparametric}
        \Expect R(\fkNN_{n,k}) - R^*
        \le \pth{C_1+C_\alpha \cdot 5^{1+\alpha}}n^{-\frac{(1+\alpha)\delta}{1+\alpha+2\delta}} + 3\exp\pth{-\frac{1}{25}n^{\frac{2\delta}{1+\alpha+2\delta}}}. 
    \end{equation}
\end{corollary}

\prettyref{cor:nonparametric} recovers the existing convergence rates $\cO(n^{-\frac{\beta(1+\alpha)}{2\beta+d}})$ by setting $\delta=\beta(1+\alpha)/d$ in either~\eqref{eq:psi-Holder} or~\eqref{eq:psi-general-Lip}. Additionally, \prettyref{cor:nonparametric} shows that a faster rate can be achieved if the power $\delta$ is large. However, existing conditions typically yield $\delta=\cO(1/d)$, which results in the rate~\eqref{eq:ub-nonparametric} being $n^{-\cO(1/d)}$. Consequently, to achieve a diminishing expected excess risk, $n$ must grow exponentially with the dimension $d$.

From our general bounds in~\prettyref{sec:general}, with a sample size $n$ that grows exponentially with $d$, a diverging neighbor order $k$ can be chosen such that $\rho=k/n$ is sufficiently small, yielding a diminishing $\psi(\rho,\sqrt{\rho})\le C_1 \rho^{\cO(1/d)}$ based on~\eqref{eq:psi-nonparametric}. Nonetheless, in scenarios with a smaller sample size, where $n$ scales only polynomially with $d$, $\psi$ could still be negligible at a larger $\rho$ if the underlying distribution exhibits favorable properties, as will be investigated in the next subsection.








\subsection{Parametric Rates}
\label{sec:parametric}
Our general bounds yield fast rates when $\psi$ is negligible at a large $\rho$. In this subsection, we consider a constant $\rho=c_0$ and leverage the value of $\psi(c_0,t)$ when $t$ is sufficiently small. For instance, under the logistic regression model~\eqref{eq:model-logit}, we will show in \prettyref{sec:logit} that the distribution in high dimensions satisfies~\eqref{eq:majorizing} with
\begin{equation}
\label{eq:psi-logit}
\psi\pth{c_0,\sqrt{c/d}} \le \cO(d/n),
\end{equation}
where $c_0,c$ are constants. 

To present the general parametric rates, let 
\begin{equation}
    \label{eq:psi-parametric}
    \psi(c_0,t)
    =
    \begin{cases}
    \epsilon,& t\le t_0,\\
    1,& t>t_0.
    \end{cases}
\end{equation}
The next result is derived by applying~\prettyref{thm:main-margin} with $t=t_0$ and $\rho=c_0$.


\begin{corollary}
    \label{cor:parametric}
    If $\psi$ specified in~\eqref{eq:psi-parametric} is a majorizing function satisfying~\eqref{eq:majorizing}, the margin condition~\eqref{eq:margin} holds, and $k=\Floor{c_0 n}\ge 5$, then   
    \begin{equation}
        \label{eq:ub-parametric}
        \Expect R(\fkNN_{n,k}) - R^*
        \le \epsilon + C_\alpha  \pth{\frac{25 }{ n t_0^2}}^{\frac{1+\alpha}{2}} + 3 e^{-\frac{c_0 n}{25}}.
    \end{equation}
\end{corollary}

\prettyref{cor:parametric} provides parametric rates when $\epsilon$ is negligible and $t_0$ decreases polynomially with $d$. Although~\eqref{eq:ub-parametric} only establishes the expected excess risk with a prescribed $t=t_0$, the choice of $t_0$ can depend on other parameters. Specifically, for a given sequence $\rho=\rho_n$, we define the \emph{critical threshold} as
\begin{equation}
\label{eq:critical}
t_n^* \triangleq 
\inf\sth{t: \psi(\rho_n,t) \ge 1/(nt^2)^{\frac{1+\alpha}{2}}}.
\end{equation}
Employing this notion, \eqref{eq:ub-parametric} simplifies to $\cO(\frac{1}{ n (t_n^*)^2})^{\frac{1+\alpha}{2}}$, with $(t_n^*)^{-2}$ characterizing the \emph{effective dimension} of the distribution. Consequently, our results offer a theoretical underpinning for the fast rate observed in \prettyref{fig:syn}, demonstrating that $(t_n^*)^{-2}\le \cO(d)$ in view of~\eqref{eq:psi-logit}.



\subsection{Multiphase Rates}
\label{sec:multiphase}

We note that the majorization $\psi(\rho, t)$ is an upper bound as defined in~\eqref{eq:majorizing}. Therefore, the conditions~\eqref{eq:psi-nonparametric} and~\eqref{eq:psi-parametric} for nonparametric and parametric rates are not contradictory. More specifically, it is possible that for some constant $ \epsilon \in (0, 1)$ we have,

\begin{equation}\label{eq:psi-two-phase}
    \psi(\rho, t) \le \epsilon \wedge \rho^{1/d}.
\end{equation}
\begin{corollary}
    \label{cor:multiphase}
    Let $\psi$ be the function specified in~\eqref{eq:psi-two-phase} and $k=\Floor{n^{3/4}}$, where $\rho = \tfrac{1}{\sqrt[4]{n}} \le t_0$.
    Then,
    \begin{equation}
        \label{eq:ub-two-phase}
        \Expect R(\fkNN_{n,k}) - R^*
        \lesssim  \min \left\{ \epsilon + \pth{\frac{1}{ n t_0^2}}^{\frac{1+\alpha}{2}}, \ n^{-(\frac{\delta}{4} \wedge \frac{3+3\alpha}{8} )} \right\} .
    \end{equation}
\end{corollary}

We note that the above corollary gives a two-phase bound. When the excess risk is greater than $\epsilon$, the decay is polynomially fast in the sample size $n$. When the excess risk is below $\epsilon$, the learning process suffers from the curse of the dimension when $\delta = \calO(1/d)$. 

The above two phase rates happens when data distribution is nonlinear and inhomogeneous. For some data $x$, fast rates can be achieved, whereas for other data $x'$, the curse of dimension happens. 
The two-phase behavior is in fact more common than a uniform convergence rate, as illustrated in Figure~\ref{fig:mnist}.
We will see in Section~\ref{sec:two-phase} that the two-phase rate provably happen for a slightly modified logistic model with a linear decision boundary.

\subsection{Minimax Lower Bounds}
The minimax risk over a family of distributions is another commonly used risk measure. While our general bounds yield distribution-dependent expected excess risk, they also provide minimax upper bounds when $\psi$ serves as the envelope majorizing function. This subsection complements the error rates by establishing corresponding minimax lower bounds for specific distribution families. 

For a given function $\psi:(0,1)\times \reals \to [0,1]$, let $\calP(\psi)$ denote the set of all joint distributions of $(X,Y)\in \reals^d\times \{0,1\}$ with $d\in\naturals$ that satisfy~\eqref{eq:majorizing}. Note that the definition of $\calP(\psi)$ depends only on the function $\psi$ and allows distributions in any finite-dimensional spaces. This setup accommodates scenarios where the intrinsic dimension, rather than the ambient dimension, determines the convergence rates~\cite{kpotufe2011,gottlieb2016nearly}. Let $\calM_\alpha$ denote the set of joint distributions of $(X,Y)$ that satisfy~\eqref{eq:margin}. Further, Let $\calP(\psi,\alpha)\triangleq \calP(\psi) \cap \calM_\alpha$.

\begin{theorem}
    \label{thm:lb-nonparametric}
    Let $\psi$ be the function specified in~\eqref{eq:psi-nonparametric}.
    If $\delta$ and $\alpha$ are constants satisfying $\delta \le \frac{1+\alpha}{\alpha}$, then 
    \begin{equation}
        \label{eq:lb-nonparametric}
        \inf_{\hat f}\sup_{P\in \calP(\psi,\alpha)} \Expect R(\hat f) - R^*
        \gtrsim n^{-\frac{(1+\alpha)\delta}{1+\alpha+2\delta}}.
    \end{equation}
\end{theorem}

\begin{theorem}
    \label{thm:lb-parametric}
    Let $\psi$ be the function specified in~\eqref{eq:psi-parametric}.
    If $\alpha=1$, $c_0\le 1/12$, $t_0 \lesssim 1/\sqrt{\log n}$, and $\epsilon\lesssim 1/(n t_0^2)$, then
    \begin{equation}
        \label{eq:minimax-parametric}
        \inf_{\hat f}\sup_{P\in \calP(\psi,1)} \Expect R(\hat f) - R^*(P)
        \gtrsim  \frac{1}{n t_0^2}.
    \end{equation}
\end{theorem}

We derive lower bounds for the minimax risk by reducing to appropriate subsets of distributions with known minimax lower bounds. For~\prettyref{thm:lb-nonparametric}, we consider the family of distributions satisfying $\beta$-H\"older continuity~\eqref{eq:smoothness} and the strong density condition; for~\prettyref{thm:lb-parametric} we consider the family of standard logistic regression models. 

\subsection{Proofs}
\label{sec:pf-rates}

We first prove the validity of the majorizing functions under conditions in the existing literature. 
\begin{prop}
    \label{prop:strong-density}
    Suppose a distribution $P$ satisfies \eqref{eq:smoothness}, \eqref{eq:margin}, and $\mu_r(x)\ge \kappa r^d$.
    Then $P\in \calP(\psi)$ with 
    \[
    \psi(\rho,\sqrt{\rho}) = C_0 \pth{2L \pth{2/\kappa}^{\frac{\beta}{d}}}^{1+\alpha} \cdot \rho^{\frac{\beta(1+\alpha)}{d}}.
    \]
\end{prop}
\begin{proof}
    Consider a point $x$ satisfying $\eta(x)>\frac{1}{2}$ without loss of generality. 
    The relative signal strength for a fixed radius $r$ is lower bounded by
    \begin{align*}
        \frac{\pi_1 \mu_{1,r}(x)}{\pi_0 \mu_{0,r}(x)}-1
        & = \frac{\int_{B_r(x)}\eta(u)P_X(du)}{\int_{B_r(x)}(1-\eta(u))P_X(du)} -1 \\
        & = \frac{\int_{B_r(x)}(2\eta(u)-1)P_X(du)}{\int_{B_r(x)}(1-\eta(u))P_X(du)} \\
        & \ge \frac{\int_{B_r(x)}(2\eta(u)-1- 2L r^{\beta})P_X(du)}{\int_{B_r(x)}(1-\eta(u))P_X(du)},
    \end{align*}
    where the last step is according to the $\beta$-H\"older continuity~\eqref{eq:smoothness}.
    For any radius $r$ satisfying $\frac{\rho}{2}\le \mu_r(x)\le 2\rho$, by the strong density assumption~\eqref{eq:sda}, it necessarily holds that $\kappa r^d \le 2\rho$, which is equivalent to
    \[
    r \le \pth{\frac{2\rho}{\kappa}}^{\frac{1}{d}}.
    \]
    Therefore, if $2\eta(x)-1> 4 L (\frac{2\rho}{\kappa})^{\beta/d}\ge 4Lr^\beta$, then
    \[
    \sqrt{\rho}\pth{\frac{\pi_1 \mu_{1,r}(x)}{\pi_0 \mu_{0,r}(x)}-1}
    > \sqrt{\rho}\frac{(\eta(x)-\frac{1}{2})\int_{B_r(x)}P_X(du)}{\int_{B_r(x)}(1-\eta(u))P_X(du)}
    \ge \sqrt{\rho}\abs{\eta(x)-\frac{1}{2}},
    \]
    where the last inequality applied $0\le 1-\eta(u)\le 1$.
    We obtain that
    \[
    \tau_\rho(x)\ge \sqrt{\rho}\abs{\eta(x)-\frac{1}{2}}.
    \]
    Equivalently, the contrapositive of the above analysis is
    \[
    E_\rho(t)
    \subseteq \sth{x: \abs{\eta(x)-\frac{1}{2}}\le 2L \pth{\frac{2\rho}{\kappa}}^{\frac{\beta}{d}}},
    \]
    where $t=\sqrt{\rho}$.
    Applying the margin condition~\eqref{eq:margin} yields 
    \[
    \sup_{x\in E_\rho(t)} |2\eta(x)-1| \cdot P_X(E_\rho(t)) 
    \le C_0 \pth{2L \pth{\frac{2\rho}{\kappa}}^{\frac{\beta}{d}}}^{1+\alpha}.
    \]
    Consequently, the right-hand side of the above is a valid majorizing function satisfying~\eqref{eq:majorizing}.
\end{proof}

\begin{prop}
    \label{prop:generalized-Lipschitz}
    Suppose a distribution $P$ satisfies \eqref{eq:margin}, and there is an increasing function $h$ with $\lim_{s\downarrow 0}h(s)=0$ such that $|\eta(x)-\eta(y)|\le h(P_X(B_{\Norm{x-y}_2}(x))$.
    Then $P\in \calP(\psi)$ with 
    \[
    \psi(\rho,\sqrt{\rho}) = C_0 \pth{4 h(2\rho)}^{1+\alpha}.
    \]
\end{prop}
\begin{proof}
    Similar to the proof of~\prettyref{prop:strong-density}, consider a point $x$ satisfying $\eta(x)>\frac{1}{2}$ without loss of generality.
    For a fixed radius $r$,
    \begin{equation}
        \label{eq:relative-formula}        
        \frac{\pi_1 \mu_{1,r}(x)}{\pi_0 \mu_{0,r}(x)}-1 
        = \frac{\int_{B_r(x)}(2\eta(u)-1)P_X(du)}{\int_{B_r(x)}(1-\eta(u))P_X(du)}.
    \end{equation}
    For the numerator of~\eqref{eq:relative-formula}, by the triangle inequality,
    \begin{align*}
    \int_{B_r(x)}(2\eta(u)-1)P_X(du)
    &= (2\eta(x)-1)\int_{B_r(x)}P_X(du)  + 2 \int_{B_r(x)}(\eta(u)-\eta(x))P_X(du)\\
    &\ge (2\eta(x)-1)\mu_r(x) - 2 \int_{B_r(x)} h(\mu_{\Norm{x-u}_2}(x)) P_X(du).
    \end{align*}
    Since $h$ is an increasing function, for any radius satisfying $\mu_r(x)\le 2\rho$, we have $h(\mu_{\Norm{x-u}_2}(x))\le h(\mu_{r}(x)) \le h(2\rho)$.
    Hence,
    \[
    \int_{B_r(x)}(2\eta(u)-1)P_X(du)\ge (2\eta(x)-1-2h(2\rho))\mu_r(x).
    \]
    For the denominator of~\eqref{eq:relative-formula}, since $0\le 1-\eta(u)\le 1$, we have
    \[
    0\le \int_{B_r(x)}(1-\eta(u))P_X(du)\le \mu_r(x). 
    \]
    Therefore, if $2\eta(x)-1> 4 h(2\rho)$, then 
    \[
    \sqrt{\rho}\pth{\frac{\pi_1 \mu_{1,r}(x)}{\pi_0 \mu_{0,r}(x)}-1}
    >\sqrt{\rho}\abs{\eta(x)-\frac{1}{2}}.
    \]
    Equivalently, for $t=\sqrt{\rho}$,
    \[
    E_\rho(t)
    \subseteq \sth{x: \abs{\eta(x)-\frac{1}{2}}\le 4 h(2\rho)}.
    \]
    Applying the margin condition~\eqref{eq:margin} yields 
    \[
    \sup_{x\in E_\rho(t)} |2\eta(x)-1| \cdot P_X(E_\rho(t)) 
    \le C_0 \pth{4 h(2\rho)}^{1+\alpha}.
    \]
    Consequently, the right-hand side of the above is a valid majorizing function satisfying~\eqref{eq:majorizing}.
\end{proof}

\begin{proof}[Proof of~\prettyref{thm:lb-nonparametric}]
    Consider the family of distributions $\calP_\Sigma$ over $ \reals^d\times \{0,1\}$ satisfying the $\alpha$-margin condition~\eqref{eq:margin}, the $\beta$-H\"older continuity~\eqref{eq:smoothness}, and the strong density assumption 
    \begin{equation}
        \label{eq:sda}
        \mu_r(x)\ge \kappa r^d
    \end{equation}
    as in \cite[Definition 2.2]{AT07} and \cite[Assumption A.3]{GKM16}, where the parameters $\alpha,\beta,L,\kappa,d$ are constants to be determined.   
    Applying the lower bound of $\tau_\rho(x)$ in \prettyref{prop:strong-density},
    by taking the parameters $\beta$ and $d$ of $\calP_\Sigma$ satisfying $\frac{\beta}{d}=\frac{\delta }{1+\alpha}$ and $C_0 \pth{2L \pth{\frac{2}{\kappa}}^{\frac{\beta}{d}}}^{1+\alpha}\le C_1$, $\calP_\Sigma$ is a subset of $\calP(\psi,\alpha)$ for the function $\psi$ specified in~\eqref{eq:psi-parametric}, and it follows that 
    \[
    \inf_{\hat f}\sup_{P\in \calP(\psi,\alpha)} \Expect R(\hat f) - R^*(P)
    \ge \inf_{\hat f}\sup_{P\in \calP_\Sigma} \Expect R(\hat f) - R^*(P).
    \]
    By the assumption $\delta \le \frac{1+\alpha}{\alpha}$, we have $\alpha \beta = \frac{\delta \alpha }{1+\alpha}d\le d$. Then, applying the minimax lower bound in~\cite[Theorem 3.5]{AT07} yields that
    \[
    \inf_{\hat f}\sup_{P\in \calP(\psi,\alpha)} \Expect R(\hat f) - R^*(P)
    \gtrsim n^{-\frac{(1+\alpha)\delta}{1+\alpha+2\delta}}.
    \]
    The proof is completed
\end{proof}

\begin{proof}[Proof of~\prettyref{thm:lb-parametric}]
    Consider the family of distributions $\calL_d$ for the $d$-dimensional logit model with normal design as introduced in \prettyref{sec:logit}.
    For $P\in \calL_d$, it is shown in~\prettyref{prop:logit-dominance} that,
    for all $x$ with $\Norm{x}_2 \le C\sqrt{d}$,
    \[
    \frac{\tau_\rho(x)}{\abs{\eta(x)-\frac{1}{2}}} > \sqrt{\frac{c}{d}}.
    \]
    Applying Bernstein inequality as in the proof of~\prettyref{thm:logit}, we obtain
    \[
        \psi(c_0,\sqrt{c/d})\le 2e^{-C'd}.
    \]    
    Therefore, by choosing $d=\Floor{c/t_0^2}\le c/t_0^2$, we have $d\asymp t_0^{-2} \gtrsim \log n$ and thus $\psi(\rho,t)\lesssim 1/(n t_0^2)$.
    Hence, $P \in \calP(\psi)$ for $\psi$ specified in~\eqref{eq:psi-parametric}.
    Furthermore, $P\in \calM_1$ by~\prettyref{lmm:logit-margin}.
    Consequently, $\calL_d\subseteq \calP(\psi,1)$. Then, by monotonicity, 
    \[
    \inf_{\hat f}\sup_{P\in \calP(\psi,1)} \Expect R(\hat f) - R^*(P)
    \ge \inf_{\hat f}\sup_{P\in \calL_d} \Expect R(\hat f) - R^*(P).
    \]

    Next we prove the minimax lower bound for the excess risk over $\calL_d$.
    Consider $P\in \calL_d$ with $\eta(x)= \frac{1}{1+\exp(-x^\top \theta_*)}$, where $\theta_*\in S^{d-1}$.
    Recall that $P_X$ denotes the $d$-dimensional standard Gaussian measure and $f^*(x)=\indc{x^\top \theta_* >0}$.
    Given any classifier $\hat f$, define 
    \[
    \hat\theta = \argmin_{\theta\in S^{d-1}}P_X\pth{\sth{x: \hat f(x) \ne \indc{x^\top \theta >0}}}.
    \]
    Then,
    \begin{align*}
        &~P_X\pth{\sth{x: \indc{x^\top \hat\theta >0} \ne \indc{x^\top \theta_* >0}}}\\
        \le &~P_X\pth{\sth{\indc{x^\top \hat\theta >0} \ne \hat f(x) }} + P_X\pth{\sth{\indc{x^\top \theta_* >0} \ne \hat f(x) }} \\
        \le &~2P_X\pth{\sth{\indc{x^\top \theta_* >0} \ne \hat f(x) }} = 2 P_X\pth{\sth{\hat f(x) \ne f^*(x) }},
    \end{align*}
    where the second inequality follows from the optimality of $\hat \theta$.
    For the standard Gaussian measure $P_X$ and two hyperplanes $x^\top \hat\theta$ and $x^\top \theta_*$, we have
    \[
        P_X\pth{\sth{x: \indc{x^\top \hat\theta >0} \ne \indc{x^\top \theta_* >0}}}
        = \frac{|\angle (\hat \theta,\theta_*)|}{\pi} 
        \ge \frac{1}{\pi} \Norm{\hat \theta - \theta_*}_2.
    \]
    Consequently, 
    \begin{equation}
        \label{eq:estimation vs d-Delta}
        \Expect\Norm{\hat \theta - \theta_*}_2 
        \le 2\pi \int \Prob[\hat f(x) \ne f^*(x)] P_X(\diff x). 
    \end{equation}

    Recall that $M(t) = \{x: |\eta(x)-\frac{1}{2}|\le t\}$.
    For any $t>0$, we have
    \begin{align}
        &\phantom{{}={}}\int \Prob[\hat f(x) \ne f^*(x)] P_X(\diff x)\nonumber\\
        &= \pth{\int_{M(t)}+\int_{M^c(t)}}\Prob[\hat f(x) \ne f^*(x)] P_X(\diff x) \nonumber\\
        &\le P_X(M(t)) + \int_{M^c(t)} \frac{|2\eta(x)-1|}{2t}\Prob[\hat f(x) \ne f^*(x)] P_X(\diff x) \nonumber\\
        &\le C t + \frac{R(\hat f) - R^*(P)}{2t},\label{eq:d-Delta vs excess}
    \end{align}
    where the last inequality applied the upper bound $P_X(M(t))$ in~\prettyref{lmm:logit-margin}. 
    Combining~\eqref{eq:estimation vs d-Delta} and~\eqref{eq:d-Delta vs excess} and taking supremum over $P\in\calL_d$ yield that
    \[
    \sup_{P\in \calL_d} R(\hat f) - R^*(P)
    \ge 2t \pth{ \frac{1}{2\pi}\sup_{P\in \calL_d}\Expect\Norm{\hat \theta - \theta_*}_2  - Ct }.
    \]
    By applying the minimax lower bound of $\Expect\Norm{\hat \theta - \theta_*}_2$ from \cite[Theorem 1]{hsu2024sample} and picking $t\asymp \sqrt{d/n}$, we obtain
    \[
        \sup_{P\in \calL_d} R(\hat f) - R^*(P) \gtrsim \frac{d}{n}.
    \]
    Finally, since $d\asymp t_0^{-2}$, we conclude the proof of the lower bound.
\end{proof}

\section{Proofs for \prettyref{sec:fine-grained}}
For ease of notations, we define the probability and conditional probabilities within a ball $B_r(x)$ of radius $r$ centered at $x$ as
\begin{align*}
    \mu_r(x)\triangleq P_X(B_r(x)), 
    \qquad
    \mu_{i,r}(x)\triangleq P_{i}(B_r(x)) \text{ for }  i\in \{0,1\}.
\end{align*}
For clarity, we also explicitly write the Bayes risk $R^*=R^*(P)$ as a function of the underlying distribution $P$. 
For the margin condition~\eqref{eq:margin}, define a set
\[
M(t)\triangleq \sth{x: 0<\abs{\eta(x)-\frac{1}{2}}\le t}.
\]

\label{sec:pf-general}

\begin{proof}[Proof of~\prettyref{prop:pointwise}]
    By symmetry, assume $f^*(x)=1$ without loss of generality. 
    It follows from the definition of $\fkNN_{k,n}$ in~\eqref{eq:vanilla-kNN} that
    \[
    \Prob[\fkNN_{k,n}(x)\ne 1]
    = \prob{\sum_{i=1}^k Y_{(i)}(x) < k/2}.
    \]
    Conditioning on the radius of $(k+1)\Th$ nearest neighbor $R_{(k+1)}(x)=r$, 
    we apply~\cite[Theorem 1]{KR92} and obtain
    \begin{equation}
        \label{eq:kNN-conditioning}
        \sum_{i=1}^k Y_{(i)}(x) | \{R_{(k+1)}(x)=r\}
        \sim \Binom(k, \Prob[Y=1|X\in B_r(x)]).
    \end{equation}

    We first establish the concentration inequality for $R_{(k+1)}(x)$. 
    For $r>0$, the event $\{R_{(k+1)}(x)\ge r\}$ is equivalent to $\{|\{ X_1,\dots,X_n\} \cap B_r(x)|\le k \}$, where the cardinality $|\{ X_1,\dots,X_n\} \cap B_r(x)|$ follows the binomial distribution with $n$ trials and success probability $\mu_r(x)$.
    Hence, for a radius $r$ with $\mu_r(x)>2\rho>\frac{k}{n}$, applying the Chernoff bound (see, e.g., \cite[Theorems 4.4 and 4.5]{MU17}) yields
    \begin{align*}
        \Prob[R_{(k+1)}(x)\ge r]
        & = \Prob[ \Binom(n, \mu_r(x))|\le k ]\\
        & \le \exp\pth{-\frac{1}{2}n \mu_r(x) \pth{1-\frac{\Floor{n\rho}}{n\mu_r(x)}}^2} \\
        & \le \exp\pth{-\frac{n\rho}{4}}.
    \end{align*}
    Similarly, for a radius $r$ with $\mu_r(x)<\frac{\rho}{2}<\frac{k}{n}$, we have
    \begin{align*}
        \Prob[R_{(k+1)}(x)< r]
        & = \Prob[\Binom(n, \mu_r(x))> k]\\
        & \le \exp\pth{-\frac{1}{3}n \mu_r(x) \pth{\frac{\Floor{n\rho}}{n\mu_r(x)}-1}^2} \\
        & \le \exp\pth{-\frac{2}{27}n\rho},
    \end{align*}
    where the last inequality used $\Floor{x}\ge \frac{5}{6}x$ when $x\ge 5$.
    Let $\calR \triangleq \{r: \frac{\rho}{2} \le \mu_r(x) \le 2\rho \}$. Therefore,
    \begin{equation}
        \label{eq:radius-tail}
        \Prob[R_{(k+1)}(x)\not\in \calR] \le 2\exp\pth{-\frac{2}{27}n\rho}.
    \end{equation}

    Next, we upper bound the probability of $\{\fkNN_{k,n}(x)\ne 1\}$ conditioning on the radius $R_{(k+1)}(x)=r\in \calR$. 
    If $\tau_\rho(x)>0$, then $\pi_1 \mu_{1,r}(x)>\pi_0 \mu_{0,r}(x)$ for all $r\in\calR$ by the definition in~\eqref{eq:def-tau}. 
    Consequently, the success probability in~\eqref{eq:kNN-conditioning} satisfies
    \[
        \Prob[Y=1|X\in B_r(x)]
        = \frac{\pi_1 \mu_{1,r}(x)}{\pi_1 \mu_{1,r}(x)+\pi_0 \mu_{0,r}(x)}
        > \frac{1}{2}.
    \]
    Hence, it follows from~\cite{Hoeffding94} that
    \begin{align*}
        \Prob[\fkNN_{k,n}(x)\ne 1 | R_{(k+1)}(x)=r] 
        & \le \pth{\sqrt{\frac{ 2 \pi_1 \mu_{1,r}(x)}{\pi_1 \mu_{1,r}(x)+\pi_0 \mu_{0,r}(x)}} \sqrt{\frac{2 \pi_0 \mu_{0,r}(x)}{\pi_1 \mu_{1,r}(x)+\pi_0 \mu_{0,r}(x)} } }^{k} \\
        & = \pth{\frac{\pth{\frac{\pi_1 \mu_{1,r}(x)}{\pi_0 \mu_{0,r}(x)}}^{\frac{1}{2}}+\pth{\frac{\pi_1 \mu_{1,r}(x)}{\pi_0 \mu_{0,r}(x)}}^{-\frac{1}{2}}}{2}  }^{-k}.
    \end{align*}
    We consider the following two cases separately. 
    \begin{itemize}
        \item If $0<\tau_\rho^2(x)\le \rho$, then $1< \frac{\pi_1 \mu_{1,r}(x)}{\pi_0 \mu_{0,r}(x)}\le 2$. We get
        \begin{align*}
        \pth{\frac{\pth{\frac{\pi_1 \mu_{1,r}(x)}{\pi_0 \mu_{0,r}(x)}}^{\frac{1}{2}}+\pth{\frac{\pi_1 \mu_{1,r}(x)}{\pi_0 \mu_{0,r}(x)}}^{-\frac{1}{2}}}{2}  }^{-k} 
        & \le \exp\pth{- \frac{k}{20} \pth{\frac{\pi_1 \mu_{1,r}(x)}{\pi_0 \mu_{0,r}(x)}-1}^2} \\
        & \le \exp\pth{-\frac{1}{24}n\tau_\rho^2(x)}
        \end{align*}
        where the first inequality used the fact that $\log\pth{\frac{1}{2}\pth{\sqrt{x}+\frac{1}{\sqrt{x}}} } \ge \frac{1}{20}(x-1)^2$ for $1<x\le 2$, and the second inequality used $k=\Floor{n\rho}\ge \frac{5}{6}n\rho$ for $k\ge 5$ and the definition of $\tau_\rho$ in~\eqref{eq:def-tau}.
        \item If $\tau_\rho^2(x)\ge \rho$, then $\frac{\pi_1 \mu_{1,r}(x)}{\pi_0 \mu_{0,r}(x)}\ge 2$. Since $x\mapsto \sqrt{x}+\frac{1}{\sqrt{x}}$ is monotonically increasing for $x\ge 1$, we get
        \begin{align*}
        \pth{\frac{\pth{\frac{\pi_1 \mu_{1,r}(x)}{\pi_0 \mu_{0,r}(x)}}^{\frac{1}{2}}+\pth{\frac{\pi_1 \mu_{1,r}(x)}{\pi_0 \mu_{0,r}(x)}}^{-\frac{1}{2}}}{2}  }^{-k} 
        & \le \pth{\frac{\sqrt{2}+\frac{1}{\sqrt{2}}}{2}}^{-k} 
        \le \exp\pth{-\frac{1}{24}n\rho},
        \end{align*}
        where the last inequality used $k\ge \frac{5}{6}n\rho$.
    \end{itemize}
    Since the upper bounds hold uniformly for all $r\in \calR$, we conclude that
    \begin{equation}
        \label{eq:err-typical}
        \Prob[\fkNN_{k,n}(x)\ne 1 | R_{(k+1)}(x)\in\calR]
        \le \exp\pth{-\frac{1}{24}n(\tau_\rho^2(x) \wedge \rho)}.
    \end{equation}

    The conclusion follows by combining \eqref{eq:radius-tail} and \eqref{eq:err-typical}. 
\end{proof}

\begin{proof}[Proof of~\prettyref{thm:main}]
    The excess risk of a classifier $f$ compared with the Bayes risk is
    \[
    R(f) - R^*(P)
    =\int |2\eta(x)-1| \indc{f(x)\ne f^*(x)} P_X(\diff x).
    \]    
    Therefore, by applying Fubini's theorem, the expected excess risk of $\fkNN_{n,k}$ is 
    \begin{equation}
    \label{eq:excess-integral-pf}
    \Expect R(\fkNN_{n,k}) - R^*(P)
    =\int |2\eta(x)-1| \Prob[\fkNN_{n,k}(x)\ne f^*(x)] P_X(\diff x).
    \end{equation}

    For any $t>0$, we have
    \begin{align}
        &~\Expect R(\fkNN_{n,k}) - R^*(P)\nonumber\\
        =&~ \pth{\int_{E_\rho(t)} + \int_{E_\rho^c(t)}} |2\eta(x)-1| \Prob[\fkNN_{n,k}(x)\ne f^*(x)] P_X(\diff x)\\
        \le&~ \psi(\rho,t) + 3\int_{E_\rho^c(t)} |2\eta(x)-1| \exp\pth{- \frac{1}{25}n(\tau_\rho^2(x)\wedge \rho) } P_X(\diff x)\nonumber\\
        \le&~ \psi(\rho,t) 
        + 6 \int \abs{\eta(x)-\frac{1}{2}} e^{-\frac{n t^2 }{25} (\eta(x)-\frac{1}{2})^2} P_X(\diff x)
        + 3 e^{-\frac{n\rho}{25}}, \label{eq:ub-general}
    \end{align}
    where the first inequality applied the majorizing property of $\psi$ in~\eqref{eq:majorizing} and~\prettyref{prop:pointwise}, and the second inequality applied $\tau_\rho(x)\ge t |\eta(x)-\frac{1}{2}|$ for $x\not\in E_\rho(t)$ and $|2\eta(x)-1|\le 1$. 
    Note that $x \exp(- a x^2) \le \frac{1}{\sqrt{2 a e}}$ for all $a,x\ge 0$. 
    It follows from~\eqref{eq:ub-general} that
    \begin{equation}
        \label{eq:ub-lambda}
        \Expect R(\fkNN_{n,k}) - R^*(P)
        \le \psi(\rho,t) + \frac{30}{\sqrt{2e n t^2}} + 3 e^{-\frac{n\rho}{25}}. 
    \end{equation}
    We conclude the proof by minimizing the right-hand side of~\eqref{eq:ub-lambda} over $t>0$.
\end{proof}

\begin{proof}[Proof of~\prettyref{thm:main-margin}]
    The analyses are similar to the proof of~\prettyref{thm:main}.
    Under the additional margin condition~\prettyref{eq:margin}, we upper bound the second term of~\eqref{eq:ub-general} using the peeling device.
    Define
    \begin{align*}
        & A_i\triangleq \sth{x: 0 <\abs{\eta(x)-\frac{1}{2}}\le \frac{5\cdot 2^{i}}{\sqrt{n t^2}} }, \quad i\ge 0,\\
        & B_i\triangleq A_{i+1} \backslash A_i=\sth{x: \frac{5\cdot 2^{i}}{\sqrt{n t^2}} <|\eta(x)-\frac{1}{2}|\le \frac{5\cdot 2^{i+1}}{\sqrt{n t^2}}},\quad i\ge 0.
    \end{align*}
    Additionally, recall the constant $t^*$ in~\prettyref{eq:margin}, and define $i^*\triangleq \max\{i\in\naturals: \frac{5\cdot 2^{i}}{\sqrt{n t^2}}\le t^*\}$. 
    It follows from the definition that $\frac{5\cdot 2^{i^*+1}}{\sqrt{n t^2}}> t^*$, which implies that $2^{i^*}\ge \frac{t^*}{10} \sqrt{n t^2} $. 
    Then,
    \begin{align*}
        &~\int \abs{\eta(x)-\frac{1}{2}} \exp\pth{-\frac{n t^2 }{25} (\eta(x)-\frac{1}{2})^2} P_X(\diff x)\\
        \le &~ \pth{\int_{A_0}+\sum_{i=0}^{i^*-1} \int_{B_i} + \int_{A_{i^*}^c} }\abs{\eta(x)-\frac{1}{2}} \exp\pth{-\frac{n t^2 }{25}(\eta(x)-\frac{1}{2})^2} P_X(\diff x) \\
        \le &~\frac{5}{\sqrt{n t^2}} P_X(A_0) + \sum_{i=0}^{i^*-1} \pth{\frac{5\cdot 2^{i}}{\sqrt{n t^2}} \exp\pth{-4^i} P_X(B_i)} + \frac{\exp\pth{-4^{i^*}}}{2} P(A_{i^*}^c)\\ 
        = &~(1-e^{-1})\frac{5}{\sqrt{n t^2}} P_X(A_0) + \frac{5}{\sqrt{n t^2}}\sum_{i=1}^{i^*-1} \pth{2^{i-1} \exp\pth{-4^{i-1}} - 2^{i} \exp\pth{-4^i}} P_X(A_i)\\
        & \qquad +  \frac{5\cdot 2^{i^*-1} \exp\pth{-4^{i^*-1}}}{\sqrt{n t^2}} P(A_{i^*}) + \frac{\exp\pth{-4^{i^*}}}{2} P(A_{i^*}^c),
    \end{align*}
    where in the second inequality we applied the fact that the function $x\mapsto xe^{-a x^2}$ is monotone decreasing on $x \ge \frac{1}{\sqrt{2a}} $.
    The last two terms are upper bounded as 
    \[
    \frac{5\cdot 2^{i^*-1} \exp\pth{-4^{i^*-1}}}{\sqrt{n t^2}} P(A_{i^*}) + \frac{\exp\pth{-4^{i^*}}}{2} P(A_{i^*}^c)
    \le \frac{\exp\pth{-\frac{1}{4}(\frac{t^*}{10})^2nt^2}}{2}
    \le C_\alpha' \pth{\frac{5}{\sqrt{n t^2}}}^{1+\alpha},
    \]
    where $C_\alpha'$ only depends on $\alpha$, and the last inequality holds since $\sup_{x>0} x^{\frac{1+\alpha}{2}}e^{-x}<\infty$. 
    By applying the margin condition~\prettyref{eq:margin}, we obtain 
    \begin{align*}
        &~\int \abs{\eta(x)-\frac{1}{2}} \exp\pth{-\frac{n t^2}{25}(\eta(x)-\frac{1}{2})^2} P_X(\diff x)\\
        \le &~  \pth{\frac{5}{\sqrt{n t^2}}}^{1+\alpha}\pth{C_0 + C_0\sum_{i=1}^{i^*-1} 2^{i\alpha}\pth{2^{i-1} \exp\pth{-4^{i-1}} - 2^{i} \exp\pth{-4^i}} + C_\alpha'}\\
        = &~ \pth{\frac{5}{\sqrt{n t^2}}}^{1+\alpha}\pth{C_0 + C_0\sum_{i=1}^{i^*-1} 2^{i\alpha} \int \indc{2^{i-1}< x \le 2^i} d(-xe^{-x^2})  + C_\alpha'}\\
        \le &~ \pth{\frac{5}{\sqrt{n t^2}}}^{1+\alpha} \pth{C_0 + C_0\int_1^\infty (2x)^{\alpha} d(-xe^{-x^2})  + C_\alpha'}\\
        = &~ \pth{\frac{5}{\sqrt{n t^2}}}^{1+\alpha} \pth{C_0 +  C_0\int_1^\infty (2x)^{\alpha} (2x^2-1)e^{-x^2}\diff x  + C_\alpha'}.
    \end{align*}
    Again, it follows from~\eqref{eq:ub-general} that
    \begin{equation}
        \label{eq:ub-margin-lambda}
        \Expect R(\fkNN_{n,k}) - R^*(P)
        \le \psi(\rho,t) 
        + C_\alpha \pth{\frac{5}{\sqrt{n t^2}}}^{1+\alpha}
        + 3 e^{-\frac{n\rho}{25}}.
    \end{equation}
    where $C_\alpha \triangleq 6(C_0  +  C_0 \int_1^\infty (2x)^{\alpha} (2x^2-1)e^{-x^2}\diff x + C_\alpha')$.
    We conclude \prettyref{thm:main-margin} by minimizing the right-hand side of~\eqref{eq:ub-margin-lambda} over $t>0$.
\end{proof}

\section{Proofs for~\prettyref{sec:app}}
\subsection{Proofs for~\prettyref{sec:logit}}
\label{sec:dominance}

We verify the margin condition in \prettyref{eq:margin} for the logistic model. 

\begin{lemma}
    \label{lmm:logit-margin}
    For $P\in\calL_d$, there exists a constant $C$ such that
    \[
    P_X(M(t)) \le C t.
    \]
\end{lemma}
\begin{proof}
    It suffices to prove that $P_X(M(t)) \le C t$ for $t\le t^*$ with a constant $t^*$, 
    since $P_X(M(t))\le 1 \lesssim t^*\le t$ for $t\ge t^*$.
    The regression function satisfies
    \[
    \abs{\eta(x)-\frac{1}{2} }
    = \frac{1}{2}\abs{\frac{1-e^{-x^\top \theta_*}}{1+e^{-x^\top \theta_*}}}
    \asymp \abs{x^\top \theta_*} \wedge 1.
    \]    
    Then, the set $M(t)$ is characterized by 
    \[
    M(t) = \sth{x: |x^\top \theta_*| \wedge 1 \le C_0 t},
    \]
    where $C_0$ is a constant.
    Therefore, for $t\le \frac{1}{C_0}$, the set $M(t)$ reduces to $\{x: |x^\top \theta_*| \le C_0 t\}$.
    Since $x^\top \theta_*\sim N(0,1)$, we obtain
    \[
    \int_{-C_0t}^{-C_0t} \phi(u) \diff u
    \le \frac{2C_0t}{\sqrt{2\pi}}.
    \]
    The conclusion follows. 
\end{proof}

We will apply~\prettyref{cor:parametric} to prove the fast rate under the logistic model. To this end, it remains to verify the majorizing function in~\eqref{eq:psi-parametric} for some $t_0$ that scales with the dimension at the rate $O(\frac{1}{\sqrt{d}})$. 
The key insight is the stochastic dominance relation between two random variables representing the distances from $x$ to the covariates $X$ conditioning on $Y=1$ and $0$, respectively. 
For instance, if a point $x$ belongs to the halfspace $\{x: x^\top \theta_*>0\}$, then the following stochastic dominance condition holds (see~\prettyref{lmm:quant-dominance}):
\begin{equation}
\label{eq:dominance}
\Prob[\Norm{X-x}_2\le r | Y=1] \ge \Prob[\Norm{X-x}_2\le r | Y=0], \quad \forall r>0.
\end{equation}
The condition in particular implies that $\tau_\rho(x)\ge 0$ for all $\rho$.
\prettyref{prop:logit-dominance} provides a quantitative lower bound of $\tau_\rho(x)$ when $\rho$ is a small constant.

\begin{prop}
    \label{prop:logit-dominance}
    Suppose $\Norm{x}_2 \le C \sqrt{d}$ and $\rho\le \frac{1}{12}$ is a constant. 
    There exist a constant $c$ that only depends on $C$ and $\rho$ such that,
    \[
    \frac{\tau_\rho(x)}{\abs{\eta(x)-\frac{1}{2}}} \ge \sqrt{\frac{c }{d}} .
    \]
\end{prop}


\begin{proof}
Note that the covariate $X$ can be decomposed into independent and orthogonal components $\Pi(X)=\theta_* \theta_*^\top X$ and $\Pi_\perp(X) = (I- \theta_*\theta_*^\top)X$, and the regression function $\eta(x)$ is a function of $\Pi(x)$ only. 
In the sequel, $\Pi(X)$ and $\Pi_\perp(X)$ are referred to as the signal and noise components, respectively.

Without loss of generality, we consider a point $x$ with $x^\top \theta_*>0$.
By the independence and orthogonality of $\Pi(X)$ and $\Pi_\perp(X)$, the probability $\mu_{i,r}(x)$ can be written as a convolution
\begin{align*}
    \mu_{i,r}(x)
    &=\prob{X\in B_r(x) | Y=i}\\
    &=\prob{ \Norm{\Pi_{\perp}(X-x)}_2^2  + \Norm{\Pi(X-x)}_2^2 \le r^2 \mid Y=i}\\
    &=\int_0^{r^2} f(x,r^2-u) \Prob[\Norm{\Pi(X-x)}_2^2\le u \mid Y=i] \diff u, 
\end{align*}
where $f(x,u)$ denotes the probability density function of $\Norm{\Pi_{\perp}(X-x)}^2$ supported on $\reals_+$ satisfying 
\[
    \prob{\Norm{\Pi_{\perp}(X-x)}_2^2 \le t} 
    = \int_{-\infty}^t f(x,u) \diff u.
\]
For the signal component, the cumulative distribution functions of $\Norm{\Pi(X-x)}_2$ conditioned on the label $Y$ are denoted by 
\[
\nu_{i}(x,r) \triangleq \Prob[\Norm{\Pi(X-x)}_2\le r \mid Y=i],\quad \text{for}~i \in\{0,1\}.
\]
To establish a lower bound of $\tau_\rho(x)$, we need a lower bound of
\begin{equation}
    \label{eq:relative-strength}
    \frac{\mu_{1,r}(x)}{\mu_{0,r}(x)}-1
    = \frac{\int_0^{r} f(x,r^2-u^2) \cdot (\nu_1(x,u) - \nu_0(x,u)) \cdot 2u \diff u }{\int_0^{r} f(x,r^2-u^2) \cdot \nu_0(x,u) \cdot 2u \diff u }
\end{equation}
that holds true uniformly over the typical radius $r$ satisfying $\frac{\rho}{2}\le \mu_r(x)\le 2\rho$.
To this end, we derive an upper bound of the denominator and a lower bound of the numerator.
For the denominator, by the stochastic dominance property~\eqref{eq:dominance},
we have $\mu_{0,r}(x) \le \mu_{1,r}(x)$ for all $r>0$. Then, the denominator of~\eqref{eq:relative-strength} is upper bounded by 
\begin{equation}
\label{eq:denominator}
\mu_{0,r}(x)
\le \pi_0 \mu_{0,r}(x) + \pi_1 \mu_{1,r}(x)
= \mu_r(x)
\le 2\rho.
\end{equation}
The lower bound for the numerator of~\eqref{eq:relative-strength} is based on the following two lemmas: \prettyref{lmm:quant-dominance} provides a lower bound of $\nu_1(x,u) - \nu_0(x,u)$ by quantifying the stochastic dominance property; \prettyref{lmm:gaussian-approx} is the Gaussian approximation for $f(x,r^2-u^2)$ and $\mu_r(x)$ similar to the local limit theorem and the Berry-Esseen theorem, respectively.


\begin{lemma}
    \label{lmm:quant-dominance}
    For all $r>0$, we have $\nu_1(x,r) \ge \nu_0(x,r)$ and thus $\mu_{1,r}(x) \ge \mu_{0,r}(x)$. 
    Furthermore, there exists an absolute constant $c_1$ such that, if $|r - (x^\top \theta_* \vee 1)| \le \frac{1}{4}$, then
    \[
    \nu_1(x,r)  -  \nu_0(x,r) \ge c_1 (x^\top \theta_* \wedge 1). 
    \]
\end{lemma}

For the statements of Gaussian approximation, we introduce the following notations for the mean and variance:
\begin{align*}
    &m(x)\triangleq\Expect[\Norm{\Pi_\perp(X-x)}_2^2]=d-1+\Norm{x}_2^2-(x^\top\theta_*)^2,\\
    &\tilde m(x)\triangleq\Expect[\Norm{X-x}_2^2]=d+\Norm{x}_2^2, \\
    &v^2(x)\triangleq\var[\Norm{\Pi_\perp(X-x)}_2^2]=2(d-1)+4(\Norm{x}_2^2-(x^\top\theta_*)^2),\\
    &\tilde v^2(x)\triangleq\var[\Norm{X-x}_2^2]=2d+4\Norm{x}_2^2.
\end{align*}
\begin{lemma}
    \label{lmm:gaussian-approx}
    There exist absolute constants $c_2$ and $C_2$ such that, for $d\ge C_2$, 
    \begin{align}
        &\sup_t \abs{f(x,t) - \frac{1}{v(x)}\phi\pth{\frac{t-m(x)}{v(x)}}}\le \frac{c_2}{v^2(x)},\label{eq:gaussian-PDF}\\
        &\sup_r \abs{\mu_r(x) - \Phi\pth{\frac{r^2-\tilde m(x)}{\tilde v(x)}}} \le \frac{c_2}{\tilde v(x)}.\label{eq:gaussian-CDF}
    \end{align}
\end{lemma}

We first provide a range for the typical radius by applying~\prettyref{lmm:gaussian-approx}.
When $d$ exceeds some constant such that $ \rho \ge  \frac{6 c_2}{\tilde v(x)}$, for the typical radius $r$ satisfying $\frac{\rho}{2}\le \mu_r(x) \le 2\rho$, we have
\begin{equation}
\label{eq:radius-typical}
r^2 = \Phi^{-1}(c\rho) \tilde v(x)  + \tilde m (x),
\end{equation}
where $\frac{1}{3} \le c \le 3$.
Next, we lower bound $\tau_\rho(x)$ for two cases according to the magnitude of $x^\top \theta_*$.

\paragraph{Small signal regime.}
Suppose $x^\top \theta_*\le 1$. 
By applying~\prettyref{lmm:quant-dominance}, the numerator of~\eqref{eq:relative-strength} is lower bounded by 
\begin{align}
&~\int_0^{r} f(x,r^2-u^2) \cdot (\nu_1(x,u) - \nu_0(x,u)) \cdot 2u \diff u \nonumber\\
\ge &~   \int_{3/4}^{5/4} f(x,r^2-u^2) \cdot  c (x^\top \theta_*)\cdot \frac{3}{2} \diff u \nonumber\\
\gtrsim &~ (x^\top \theta_*)\cdot  \min_{\frac{3}{4}\le u \le \frac{5}{4}} f(x,r^2-u^2) \nonumber\\
\ge &~ (x^\top \theta_*)\cdot \min_{\frac{3}{4}\le u \le \frac{5}{4}} \pth{\frac{1}{v(x)}\phi\pth{\frac{r^2-u^2-m(x)}{v(x)}} - \frac{c_2}{v^2(x)}}, \label{eq:lb-numer-1}
\end{align}
where the last step applied~\prettyref{lmm:gaussian-approx}.
It remains to lower bound $\phi(z)$ with
\begin{align}
z 
&\triangleq \frac{r^2-u^2-m(x)}{v(x)} \nonumber\\
&= z'\frac{\tilde v(x)}{v(x)} + \frac{\tilde m(x)-m(x)-u^2}{v(x)} \nonumber\\
&= z' \sqrt{1+ \frac{2+4 (x^\top \theta_*)^2}{v^2(x)}} + \frac{1+(x^\top \theta_*)^2 - u^2}{v(x)}, \label{eq:z-formula}
\end{align}
by applying the typical radius $r$ in~\eqref{eq:radius-typical}, where $\Phi(z')=c\rho$.
For $\rho\le \frac{1}{12}$ such that $c\rho \le \frac{1}{4}$, we have $z'<0$ and $|z'|\asymp 1$.
Furthermore, 
\[
\abs{z^2-z'^2} = \abs{z'^2\epsilon_1 + \epsilon_2^2 + 2z'\sqrt{1+\epsilon_1}\epsilon_2}
\lesssim \frac{1}{\sqrt{d}},
\]
where $\epsilon_1=\frac{2+4 (x^\top \theta_*)^2}{v^2(x)}$, and $\epsilon_2=\frac{1+(x^\top \theta_*)^2 - u^2}{v(x)}$.
Therefore,
\[
\phi(z)
=\phi(z')\exp\pth{-\frac{z^2-z'^2}{2}}
\ge |z'| \Phi(z')\exp\pth{-\frac{z^2-z'^2}{2}}
\gtrsim \rho,
\]
where the first inequality applied the Gaussian tail $\Phi(z')\le \frac{\phi(z')}{|z'|}$ for $z'<0$.
Since $\rho \gtrsim  \frac{1}{v(x)}$, we obtain from~\eqref{eq:lb-numer-1} that
\begin{equation}
    \label{eq:lb-numer-2}    
    \int_0^{r} f(x,r^2-u^2) \cdot (\nu_1(x,u) - \nu_0(x,u)) \cdot 2u \diff u
    \gtrsim \frac{(x^\top \theta_*) \rho }{ v(x) }.
\end{equation}
Applying~\eqref{eq:denominator} and~\eqref{eq:lb-numer-2} into~\eqref{eq:relative-strength} yields that, for all $r$ satisfying $\frac{\rho}{2}\le \mu_r(x) \le 2\rho$,
\[
\frac{\mu_{1,r}(x)}{\mu_{0,r}(x)}-1
\gtrsim \frac{x^\top \theta_*}{ v(x) }
\gtrsim \frac{x^\top \theta_*}{\sqrt{d}},
\]
where the last inequality is due to the condition $\Norm{x}_2 \le C\sqrt{d}$.

\paragraph{Large signal regime.}
Suppose $x^\top \theta_*\ge 1$. 
Here we lower bound the the numerator of~\eqref{eq:relative-strength} by 
\begin{align}
&~\int_0^{r} f(x,r^2-u^2) \cdot (\nu_1(x,u) - \nu_0(x,u)) \cdot 2u \diff u \nonumber\\
\ge &~   \int_{x^\top \theta_*}^{x^\top \theta_*+ \frac{1}{4} (x^\top \theta_*)^{-1}} f(x,r^2-u^2) \cdot  c \cdot (2 x^\top \theta_*) \diff u \nonumber\\
\gtrsim &~  \min_{x^\top \theta_*  \le u \le x^\top \theta_*+ \frac{1}{4} (x^\top \theta_*)^{-1}} f(x,r^2-u^2)  \nonumber\\
\ge &~ \min_{x^\top \theta_*  \le u \le x^\top \theta_*+ \frac{1}{4} (x^\top \theta_*)^{-1}} \pth{\frac{1}{v(x)}\phi\pth{\frac{r^2-u^2-m(x)}{v(x)}} - \frac{c_2}{v^2(x)}}, \label{eq:lb-numer-large-1}
\end{align}
where the last step applied~\prettyref{lmm:gaussian-approx}.
To analyze $\phi(z)$ with the same $z$ defined in~\eqref{eq:z-formula}, we observe that $\frac{\tilde v(x)}{v(x)}\lesssim 1$ since $\Norm{x}_2\le C \sqrt{d}$, and
\[
\abs{\frac{1+(x^\top \theta_*)^2 - u^2}{v(x)}}
\lesssim \frac{1}{v(x)}
\]
for $x^\top \theta_*  \le u \le x^\top \theta_*+ \frac{1}{4} (x^\top \theta_*)^{-1}$.
Hence, $|z^2-z'^2| \lesssim 1$, and thus $\phi(z)\gtrsim \rho$ following similar arguments.
Since $\rho \gtrsim  \frac{1}{v(x)}$, we obtain from~\eqref{eq:lb-numer-large-1} that
\begin{equation}
    \label{eq:lb-numer-large-2}    
    \int_0^{r} f(x,r^2-u^2) \cdot (\nu_1(x,u) - \nu_0(x,u)) \cdot 2u \diff u
    \gtrsim \frac{\rho }{ v(x) }.
\end{equation}
Applying~\eqref{eq:denominator} and~\eqref{eq:lb-numer-large-2} into~\eqref{eq:relative-strength} yields that, for all $r$ satisfying $\frac{\rho}{2}\le \mu_r(x) \le 2\rho$, 
\[
\frac{\mu_{1,r}(x)}{\mu_{0,r}(x)}-1
\gtrsim \frac{1}{ v(x) }
\gtrsim \frac{1}{\sqrt{d}},
\]
where the last inequality is due to the condition $\Norm{x}_2 \le C\sqrt{d}$.

Combining the previous two cases, we conclude that 
\[
\frac{\mu_{1,r}(x)}{\mu_{0,r}(x)}-1
\gtrsim \frac{(x^\top \theta_*)\wedge 1}{\sqrt{d}},
\]
for all radius $r$ satisfying $\frac{\rho}{2}\le \mu_r(x) \le 2\rho$.
Note that $|\eta(x)-\frac{1}{2}|\asymp (x^\top \theta_*)\wedge 1$.
We apply the definition of $\tau_\rho(x)$ in~\eqref{eq:def-tau} and conclude the proof. 
\end{proof}

\begin{proof}[Proof of~\prettyref{thm:logit}]
    Let $\rho=c_0 < \frac{1}{12}$ be a constant. 
    According to~\prettyref{prop:logit-dominance}, there exists another constant $c'$ such that, for $t=\sqrt{c'/d}$,
    \[
    E_\rho(t)
    =\sth{x: \frac{\tau_\rho(x)}{\abs{\eta(x)-\frac{1}{2}}} \le t} 
    \subseteq \sth{x: \Norm{x}_2 \ge C\sqrt{d}}.
    \]
    Therefore, applying the Bernstein inequality yields that (see, e.g., \cite[Theorem~3.1.1]{Vershynin2018})
    \[
    P_X(E_\rho(t)) \le 2 e^{-C'd},
    \]
    where $C'$ is an absolute constant. 
    Furthermore, it is proved in~\prettyref{lmm:logit-margin} that the logistic regression model $P\in\calL_d$ satisfies the margin condition~\eqref{eq:margin} with $\alpha=1$.
    Therefore, $\calL_d \subseteq P(\psi, 1)$ with 
    \[
    \psi(\rho,t) = 2 e^{-C'd}\le \frac{d}{n},
    \]
    where the last inequality holds since $d \ge C\log n$.
    The upper bound follows by applying~\prettyref{cor:parametric}.
    The lower bound is proved in \prettyref{thm:lb-parametric}.
\end{proof}

\begin{proof}[Proof of~\prettyref{lmm:quant-dominance}]
By the definition of $\nu_i$ functions, we have 
\begin{align*}
\nu_{1}(x,r) 
&= \Prob[\Norm{\Pi(X-x)}_2\le r \mid Y=1]\\
&=\prob{X^\top \theta_* \in [x^\top\theta_*-r,x^\top\theta_*+r]  \mid Y=1 }\\
&=\int_{x^\top \theta_*-r}^{x^\top \theta_*+r} 2\phi(u)g(u) \diff u,
\end{align*}
where $g(x)=\frac{1}{1+e^{-x}}$. Similarly, $\nu_{0}(x,r) =\int_{x^\top \theta_*-r}^{x^\top \theta_*+r} 2\phi(u)g(-u) \diff u$.
Hence,
\begin{align}
\nu_{1}(x,r) - \nu_{0}(x,r) 
&= \int_{x^\top \theta_*-r}^{x^\top \theta_*+r} 2\phi(u)(g(u) - g(-u)) \diff u\nonumber\\
&= \int_{|x^\top \theta_*-r|}^{x^\top \theta_*+r} 2\phi(u)(g(u) - g(-u)) \diff u,\label{eq:v1-v0}
\end{align}
where the second equality holds since $x^\top \theta_*>0$ and $u\mapsto 2\phi(u)(g(u) - g(-u))$ is an odd function. 
Since $g(u)\ge g(-u)$ for $u\ge 0$, we conclude that $\nu_{1}(x,r) \ge \nu_{0}(x,r) $. 

For the quantitative lower bounds, we consider two cases separately.
\begin{itemize}
    \item Case 1: $x^\top \theta_* \ge 1$. 
    We establish a uniform lower bound of~\eqref{eq:v1-v0} for $r\in [x^\top \theta_*-\frac{1}{4},x^\top \theta_*+\frac{1}{4}]$. 
    In this case, the upper limit $x^\top \theta_*+r\ge 2x^\top \theta_*-\frac{1}{4}\ge \frac{7}{4}$, and the lower limit $|x^\top \theta_*-r|\le \frac{1}{4}$.
    Hence, by applying~\eqref{eq:v1-v0},
    \[
    \nu_{1}(x,r) - \nu_{0}(x,r) 
    \ge \int_{1/4}^{7/4} 2\phi(u)(g(u) - g(-u)) \diff u 
    \gtrsim 1.
    \]
    \item Case 2: $x^\top \theta_* \le 1$. The range of $r$ of interest becomes $r\in [\frac{3}{4},\frac{5}{4}]$. In this case, the integration is over a constant region with interval length 
    \[
    x^\top \theta_*+r - |x^\top \theta_*-r|
    = 2(x^\top \theta_* \wedge r)
    \gtrsim x^\top \theta_*. 
    \]
    Then, we get from~\eqref{eq:v1-v0} that
    \[
    \nu_{1}(x,r) - \nu_{0}(x,r) 
    \ge \int_{|x^\top \theta_*-r|}^{x^\top \theta_*+r} 2\phi(u)(g(u) - g(-u)) \diff u 
    \gtrsim x^\top \theta_*. 
    \]
\end{itemize}
The proof is completed. 
\end{proof}

The proof of~\prettyref{lmm:gaussian-approx} is based on the classical Fourier inversion formula, whose applications in the $\chi^2$-type distributions are studied in \cite{Zhang2005} (with an arithmetic error corrected in \cite{Wouter16}) and \cite{Seri2015}. 
However, \cite[Theorem 1]{Zhang2005} is not directly applicable in our problem since the the distribution may have unbounded mean and thus the condition $\Delta<\frac{1}{8}$ is violated.
\begin{proof}[Proof of~\prettyref{lmm:gaussian-approx}]
Let $g$ denote the probability density function of the standardized random variable $T\triangleq\frac{\Norm{\Pi_{\perp}(X-x)}^2-m(x)}{v(x)}$.
By the change of variable formula, $f(x,t)=\frac{1}{v(x)}g(\frac{t-m(x)}{v(x)})$. 
Hence,
\begin{equation}
\label{eq:change-variable}
\sup_t \abs{f(x,t) - \frac{1}{v(x)}\phi\pth{\frac{t-m(x)}{v(x)}}}
= \frac{\sup_{t}|g(t)-\phi(t)|}{v(x)}.
\end{equation}
Applying the Fourier inversion formula yields that 
\begin{align}
&\phantom{{}={}}\sup_{t}|g(t)-\phi(t)|\nonumber\\
&\le \frac{1}{2\pi}\int |\psi(t)-e^{-t^2/2}| \diff t \nonumber \\
&\le \frac{1}{2\pi}\int_{-L}^L |\psi(t)-e^{-t^2/2}| \diff t 
+ \frac{1}{2\pi}\int_{|t|>L} e^{-t^2/2} \diff t 
+ \frac{1}{2\pi}\int_{|t|>L} |\psi(t)| \diff t,  
\label{eq:fourier-decompose}
\end{align}
where $\phi(x)\triangleq \frac{1}{\sqrt{2\pi}}e^{-\frac{x^2}{2}}$ and $\psi$ denotes the characteristic function of $T$.
By choosing $L^3=c v(x)$ for some absolute constant $c$, the upper bounds for the first two terms of~\eqref{eq:fourier-decompose} follow from \cite[Proof of Theorem 1]{Zhang2005} 
\[
\frac{1}{2\pi}\int_{-L}^L |\psi(t)-e^{-t^2/2}| \diff t 
+ \frac{1}{2\pi}\int_{|t|>L} e^{-t^2/2} \diff t 
\lesssim \frac{1}{v(x)}.
\]
For the third term of~\eqref{eq:fourier-decompose}, by setting $k=d-1$ and $\lambda = \Norm{x}_2^2-(x^\top\theta_*)^2$, 
\begin{align*}
|\psi(t)| 
&= \pth{\pth{1+\frac{4t^2}{v^2(x)}}\exp\pth{\frac{2\lambda}{k}\frac{4t^2}{v^2(x) + 4t^2}}}^{-\frac{k}{4}}\\
&\le \pth{1+\frac{2t^2}{k}}^{-\frac{k}{4}} \le \pth{\frac{2t^2}{j}}^j,\quad \forall~j\le k,
\end{align*}
where the first inequality applied $e^x \ge 1+x$ and $v^2(x)=2(k+2\lambda)$, and the second inequality applied $(1+x)^k \ge \binom{k}{j}x^j \ge (\frac{k x}{j})^j$ for $j\le k$.
By taking $j=8$, we obtain that
\[
\frac{1}{2\pi}\int_{|t|>L} |\psi(t)| \diff t
\lesssim \frac{1}{v(x)}.
\]
Then it follows from~\eqref{eq:fourier-decompose} that 
\[
\sup_{t}|g(t)-\phi(t)|\lesssim \frac{1}{v(x)},
\]
and thus the proof of~\eqref{eq:gaussian-PDF} by applying~\eqref{eq:change-variable}. 
The proof of~\eqref{eq:gaussian-CDF} follows from \cite[Eq.~(III.1)]{Seri2015} and the inequality $\binom{k}{j}\ge (\frac{k}{j})^j$.
\end{proof}

\subsection{Proofs for~\prettyref{sec:two-phase}}
\label{sec:pf-logit-modified}

\subsubsection{Proof of Theorem~\ref{thm:two-phase-upper}}
\begin{proof}[Proof of Theorem~\ref{thm:two-phase-upper}]
Since the data model is symmetric across the hyperplane defined by $x^{(2)} - 2x^{(1)} = 0$, it suffices to analyze half of the region, and the other half would be exactly the same.
Consider
\begin{align}\label{eq:calX0}
    \calX_1 
    = \sth{x \in \R^d : 2x^{(1)} \ge x^{(2)}, x^{(1)}  \in  [-1, 1], x^{(2)} \in [-4, 4]}.
\end{align}
We note that this is exactly the region where $\eta(x) \le 1/2$ and hence for any $x \in \calX_1$,  $f^*(x) = 1$.

Following~\prettyref{sec:dominance}, we set $\Pi(x)= \pth{x^{(1)}, x^{(2)}, 0, ..., 0} \in \reals^d $ and $\Pi_\perp(x) = x - \Pi(x) \in \reals^d$. We note that the key properties of $\Pi(x)$ and $\Pi_\perp(x)$ still hold, that the regression function $\eta(x)$ is a function of $\Pi(x)$ only and that the signal $\Pi(X)$ and  the noise $\Pi_\perp(X)$ are independent. 
We also adopt the same notations. Let $f(x,u)$ be the p.d.f  of the noise satisfying
\[
    \prob{\Norm{\Pi_{\perp}(X-x)}_2^2 \le t} 
    = \int_{-\infty}^t f(x,u) \diff u.
\]
Additionally, $\nu_i$ is the c.d.f of the signal defined as
\begin{align}\label{eq:nu1-nu2}
 \nu_{i}(x,r) \triangleq \Prob[\Norm{\Pi(X-x)}_2\le r \mid Y=i],\quad \text{for}~i \in\{0,1\}.   
\end{align}

\paragraph{Fast phase}
We consider the case when $n \lesssim d^2$ and hence $\rho \gtrsim d^{-1/2}$. The reason for the fast phase is the same as the parametric achieved in the logistics example. 
We show that for $t \asymp \frac{1}{d}$, we have $\psi(\sqrt[4]{1/n}, t) < 1-C$ for some absolute constant  $C \in (0, 1)$.  
To this end, we analyze 
$$\tau_\rho(x) = \inf\sth{ \sqrt{\rho}\pth{\frac{\pi_1 \mu_{1,r}(x)}{\pi_0 \mu_{0,r}(x)}-1}:  \frac{\rho}{2} \le \mu_r(x) \le 2\rho  },$$ 
on the set of points with fast rates
\begin{align}\label{eq:proof-S}
    S = \{x \in \calX_1 : \ x^{(1)} \in [-0.8, 0.8], x^{(2)} \in [-2.2, -3.8] \}.
\end{align}


By Proposition~\ref{prop:rect-dominance}, we get for $t=c/d$,
    \[
    E_\rho(t)
    =\sth{x: \frac{\tau_\rho(x)}{\abs{\eta(x)-\frac{1}{2}}} \le t} 
    \subseteq S^c,
    \]
where $S^c$ is the complement of the set $S$. We note that  $P_0(E_\rho(t)) \le 1 - P_0(S) \le 0.9$. Therefore, for $\rho = n^{-1/4}, t \lesssim \frac{1}{d}$, we have $\psi(\rho, t) < 0.45$ by~\eqref{eq:majorizing}.
Hence, by Theorem~\ref{thm:main-margin}, we can set $\alpha=1$ and get 
\[
 \Expect R(\fkNN_{n,k}) - R^*
    \le 0.45 + \tfrac{Cd}{n^{3/4}}.
\]

\paragraph{Slow phase}

When $n$ is large, we show that for $t = \frac{1}{n^{1/d}}$, we have $\psi(\frac{1}{n^{1/d}}, t) < \frac{1}{n^{1/d}}$. 

Given a fixed constant $r_0 \in (0, 1/10)$, we note that when $\rho = \sqrt[4]{\tfrac{1}{n}} \lesssim r_0^{d}$, and $d \ge 10,$
\[
\frac{\rho}{2} \le \mu_r(x) \le 2\rho  \implies r \in [\frac{r_0}{2}, 2r_0].
\]
The above describes the range of radius for our analysis. Further, we denote the set of points 
\[
S_r = \{x \in \calX_1 :  \ x^{(1)} \in [-1 + r, 1 - r], x^{(2)} \in [-4+r, 2], 2x^{(1)} \ge x^{(2)}+r \}.
\]
The set $S_r$ is the set of points that are distance $r$ away from the boundary of $\calX_0$ in the first two dimensions.
For any point $x \in  S_r$, we have $|2\eta(x) - 1| \ge r$. Consequently, we also have 
$
\tfrac{\pi_1 \mu_{1,r/2}(x)}{\pi_0 \mu_{0,r/2}(x)}-1 \ge r/2.
$
Therefore, we may conclude that for any point $x \in  S_r$, 
$\tau_\rho(x) \gtrsim \sqrt{\rho}r.$

Consequently, when $\rho = \sqrt[4]{\tfrac{1}{n}} \lesssim r_0^{d}$, and $t  \lesssim \sqrt{\rho}$, we have $E_{\rho}(t) \cap \calX_0 \subseteq \calX_0 \setminus S_{2r_0}.$ Hence, we also have by \eqref{eq:majorizing} that
\[
\psi(\rho, t) \le 2r_0 \lesssim n^{-1/4d}.
\]
The result below follows by applying Theorem~\ref{thm:main-margin} with $\alpha=1$.

    \[
     \Expect R(\fkNN_{n,k}) - R^* \le  Cn^{-1/4d}.
    \]
\end{proof}
\subsubsection{A Proposition for Proving Theorem~\ref{thm:two-phase-upper}}

\begin{prop}
    \label{prop:rect-dominance}
    Suppose $x \in S$ defined in~\eqref{eq:proof-S}, $\Norm{x}_2 \le C d$ for some absolute constant $C>0$ and $d^{-1/2} \lesssim \rho\le \frac{1}{12}$. 
    There exist a constant $c$  such that,
    \[
    \frac{\tau_\rho(x)}{\abs{\eta(x)-\frac{1}{2}}} \ge \sqrt{\frac{c\rho }{d}} .
    \]
\end{prop}
\begin{proof}
Following similar analyses to the proof of~\prettyref{prop:logit-dominance}, we have
\begin{equation}\label{eq:mu1-mu0}
    \frac{\mu_{1,r}(x)}{\mu_{0,r}(x)}-1
    = \frac{\int_0^{r} f(x,r^2-u^2) \cdot (\nu_1(x,u) - \nu_0(x,u)) \cdot 2u \diff u }{\int_0^{r} f(x,r^2-u^2) \cdot \nu_0(x,u) \cdot 2u \diff u }.
\end{equation}
For the numerator of~\eqref{eq:mu1-mu0}, as $\Pi(X)=(X^{(1)},X^{(2)})$ follows the uniform distribution over a bounded rectangle, we could analytically compute $\nu_1(x,u) - \nu_0(x,u)$. 
Recall the set 
$$S = \{x \in \calX_1 : x^{(1)} \in [-0.8, 0.8], x^{(2)} \in [-2.2, -3.8] \}.$$
For any $x\in S$ and any radius $u \in [1/20, 1/10]$, we have that the distance from the ball $B_u(x)$ to the boundary of $\calX_0$ is bounded away from zero.
Let $D_u$ denote the disk $\{t: \|\Pi(t-x)\|_2 \le u\}$.
By applying the model~\eqref{eq:logit-modified}, we have
\begin{align*}
    &2\eta(t)-1 \ge \frac{1}{10}, \qquad \forall t\in D_u,  \\
    & \nu_1(x,u) - \nu_0(x,u) = 2 \int_{D_u} (2\eta(t)-1) P_X(dt) \ge 2 \cdot \frac{\pi u^2}{8} \cdot \frac{1}{10}\ge 10^{-4}.
\end{align*}
Furthermore, by symmetry, we have that for any $u >0 $ and $x \in S$,
\begin{equation}
    \label{eq:dominance-modified}
    \nu_1(x,u) \ge \nu_0(x,u).
\end{equation}
Indeed, for any $u > 0$ and $x \in S$,if the optimal decision boundary does not pass through $D_u$, then all the points are in $\calX_1$ and $\nu_1(x,u) \ge \nu_0(x,u)$. Otherwise, we can consider the set $D_u \cap \calX_0$ where $\calX_0$ is the set of points whose optimal prediction $f^*(x) = 0$. We note that the rotational symmetry of $D_u \cap \calX_0$ across the midpoint of the optimal decision boundary within $D_u \cap \calX_0$ is always in $D_u$, by our choice of $x \in S.$

Hence, by applying the above conditions, we obtain that, for any $x \in S$ and $r \ge 1/10$,
\begin{align}\label{eq:two-phase-diff}
&~\int_0^{r} f(x,r^2-u^2) \cdot (\nu_1(x,u) - \nu_0(x,u)) \cdot 2u \diff u \nonumber\\
\ge &~   \int_{1/20}^{1/10} f(x,r^2-u^2) \cdot  10^{-5} \diff u \nonumber\\
\gtrsim &~  \min_{\frac{1}{20}\le u \le \frac{1}{10}} f(x,r^2-u^2) \nonumber\\
\ge &~ \min_{\frac{1}{20}\le u \le \frac{1}{10}} \pth{\frac{1}{v(x)}\phi\pth{\frac{r^2-u^2-m(x)}{v(x)}} - \frac{c_2}{v^2(x)}}, 
\end{align}
where the last inequality follows from Lemma~\ref{lmm:gaussian-approx}, and $m(x), v(x)$ are defined below:
\begin{align*}
    &m(x)\triangleq\Expect[\Norm{\Pi_\perp(X-x)}_2^2]=d-2+ \pth{\Norm{x}_2^2- (x^{(1)})^2 - (x^{(2)})^2},\\
    &v^2(x)\triangleq\var[\Norm{\Pi_\perp(X-x)}_2^2]=2(d-2)+4\pth{\Norm{x}_2^2 - (x^{(1)})^2 - (x^{(2)})^2 }.
\end{align*}

Recall that we aim to bound $\tau_\rho(x)$ concerning the lower bound of~\eqref{eq:mu1-mu0} over the radius $r$ satisfying $ \frac{\rho}{2} \le \mu_r(x) \le 2\rho $.
We first similarly define
\begin{align*}
    &\tilde m(x)\triangleq\Expect[\Norm{X-x}_2^2]=d+\Norm{x}_2^2, \\
    &\tilde v^2(x)\triangleq\var[\Norm{X-x}_2^2]=2d+4\Norm{x}_2^2.
\end{align*}
When $d$ exceeds some constant such that $ \rho \ge  \frac{6 c_2}{\tilde v(x)}$, the typical radius $r$ in~\eqref{eq:radius-typical} holds.
To lower bound $\phi\pth{\frac{r^2-u^2-m(x)}{v(x)}}$ in~\eqref{eq:two-phase-diff}, we have
\begin{align}
z 
&\triangleq \frac{r^2-u^2-m(x)}{v(x)} \nonumber\\
&= z'\frac{\tilde v(x)}{v(x)} + \frac{\tilde m(x)-m(x)-u^2}{v(x)} \nonumber \\
&= z' \sqrt{1+ 4\frac{1+(x^{(1)})^2 + (x^{(2)})^2}{v^2(x)}} + \frac{2+(x^{(1)})^2 + (x^{(2)})^2 - u^2}{v(x)}, \label{eq:z-formula}
\end{align}
by applying the typical radius $r^2 = z' \tilde v(x)  + \tilde m (x)$ in~\eqref{eq:radius-typical}, where $\Phi(z')=c\rho$, and $\frac{1}{3} \le c \le 3$.
For $\rho\le \frac{1}{12}$ such that $c\rho \le \frac{1}{4}$, we have $z'<0$ and $|z'|\asymp 1$.
Furthermore, we note that $(x^{(1)})^2 + (x^{(2)})^2 \le 5.$ Therefore
\[
\abs{z^2-z'^2} = \abs{z'^2\epsilon_1 + \epsilon_2^2 + 2z'\sqrt{1+\epsilon_1}\epsilon_2}
\lesssim \frac{1}{\sqrt{d}},
\]
where $\epsilon_1=4\frac{1+(x^{(1)})^2 + (x^{(2)})^2}{v^2(x)}$, and $\epsilon_2=\frac{2+(x^{(1)})^2 + (x^{(2)})^2 - u^2}{v(x)}$. 
Then we have,
\[
\phi(z)
=\phi(z')\exp\pth{-\frac{z^2-z'^2}{2}}
\ge |z'| \Phi(z')\exp\pth{-\frac{z^2-z'^2}{2}}
\gtrsim \rho,
\]
where the first inequality applied the Gaussian tail $\Phi(z')\le \frac{\phi(z')}{|z'|}$ for $z'<0$. Plug this into~\eqref{eq:two-phase-diff} and then into~\eqref{eq:mu1-mu0}, we get
\begin{align*}
    \frac{\mu_{1,r}(x)}{\mu_{0,r}(x)}-1 \gtrsim 1/\sqrt{d},
\end{align*}
where the denominator of~\eqref{eq:mu1-mu0} is at most $\mu_{0,r}(x)\le \mu_r(x) \le 2\rho$ by applying \eqref{eq:dominance-modified}.
The proposition is proved by noting that $|\eta(x) -1/2| \le \frac{1}{2}$.
\end{proof}

\subsubsection{Proof of Theorem~\ref{thm:two-phase-lower}}

\begin{proof}[Proof of Theorem~\ref{thm:two-phase-lower}]
Similar to the previous proofs, we analyze half of the region without loss of generality, and the analysis for the other half would be exactly the same. Let
\begin{align}\label{eq:calX0}
    \calX_1 = \sth{x \in \R^d :  2x^{(1)} > x^{(2)}, x^{(1)}  \in  [-1, 1], x^{(2)} \in [-4, 4]}.
\end{align}
We note that this is exactly the region when $\eta(x) > 1/2$ and hence for any $x \in \calX_1$,  $f^*(x) = 1$.

Following the proof of Proposition~\ref{prop:rect-dominance}, we recall the definitions $\Pi(x)= \pth{x^{(1)}, x^{(2)}, 0, ..., 0} \in \reals^d $, $\Pi_\perp(x) = x - \Pi(x) \in \reals^d$. Additionally,  the probability density function of the noise within radius $u$ is denoted as $f(x,u)$, and  the cumulative distribution function of the signal within radius $u$ is denoted as $\nu_i(x, u)$. We also recall the definitions of $m(x), v^2(x), \tilde{m}(x), \tilde{v}^2(x)$ from the proof of Proposition~\ref{prop:rect-dominance}.

We study the set of points with slow rates
$$S = \{x \in \calX_1 :  2x^{(1)} > x^{(2)} + 0.1 > 1.1 \}.$$
Via direct integration, a key property for any point $x \in S$ is that
\begin{align}\label{eq:moment-diff}
\Expect_{X\sim P_0}\Norm{\Pi(X-x) }^2 - \Expect_{X\sim P_1}\Norm{\Pi(X-x)}^2 \le -0.1.
\end{align}
This inequality leads to slow rate, because although the optimal prediction on $\calX_1$ is $f^*(x) = 1$, the points from the opposite class $P_0$ is in expectation closer to the point $x$. The formal statement can be found below.

\begin{lemma}
\label{lemma:lower-two-phase}
    Given $x\in S$. Assume that for some $\beta, \beta' > 0$, we have $\Norm{x}_{4}^4 \le \beta d$, and $\Norm{x}_{\infty} \le \beta'$. Assume for some absolute large constant $C$, $d > C >0$.   Then there exists an absolute constant $c$ such that
    \begin{align*}
        & \prob{ \fkNN_{k,n}(x) \ne f^*(x)}
        \ge \frac{1}{2} - 2n \exp\pth{-cd \frac{1}{\beta }\wedge \frac{1}{\beta'}  },\quad \forall k=1,\dots,n.
    \end{align*}
\end{lemma}
We note that there exists an absolute constant $\epsilon > 0$ independent from $n, d$ such that
\[
S' = \{x \in \calX_1 \ |  2x^{(1)} > x^{(2)} + 0.1 > 1.1, \Norm{x}_{4}^4 \le 2\sqrt{\log(d)} d, \Norm{x}_{\infty} \le  2\sqrt{\log(d)}\}, \quad  P_{0}\pth{S'} \ge 2\epsilon,
\]
where we used the fact that $x^{(i)}$ are i.i.d standard Gaussian for $i \ge 3$. We further note for any point $x \in S'$, we have $|2\eta(x)-1| \ge 10^{-3}$ because $S'$ is bounded away from the decision boundary.
Recall that the risk can be computed with the integration below,
\[
\Expect R(\fkNN_{n,k}) - R^*    =\int |2\eta(x)-1| \Prob[\fkNN_{n,k}(x)\ne f^*(x)] P_X(\diff x).
\]
The proof of Theorem~\ref{thm:two-phase-lower} is then complete by applying Lemma~\ref{lemma:lower-two-phase} with $\beta, \beta' \asymp \sqrt{\log(d)}$. 

\end{proof}

\subsubsection{Proof of Lemmas}
\begin{proof}[Proof of Lemma~\ref{lemma:lower-two-phase}]
Given $x\in S$ as stated in Lemma~\ref{lemma:lower-two-phase} with $f^*(x)=1$, let $R_{(k)}(x)\triangleq \Norm{X_{(k)}(x)-x}_2$ be the distance to the $k\Th$ closest point to $x$, and $R_{(n+1)}(x)\triangleq \infty$.
If $k=n$, where the classification rule reduces to the majority vote of all training data, then by symmetry,
\[
\Prob[\fkNN_{k,n}(x) \ne f^*(x)] = \frac{1}{2}.
\]
If $k\le n-1$, following similar analyses for the pointwise error in the proof of~\prettyref{prop:pointwise}, 
\[
\sum_{i=1}^k Y_{(i)}(x) | \{R_{(k+1)}(x)=r\}
        \sim \Binom(k, \Prob[Y=1|X\in B_r(x)]).
\]
Define the bad region of radii as
\[
\calR_b \triangleq \{r: \Prob[Y=1|X\in B_r(x)]< 1/2\}
=\{r: \mu_{1,r}(x)< \mu_{0,r}(x)\}.
\]
Then, by the symmetry of the binomial distribution, we get 
\[
    \Prob[\fkNN_{k,n}(x)\ne f^*(x) \mid R_{(k+1)}(x) \in \calR_b] 
    \ge \frac{1}{2}.
\]
Consequently,
\begin{align}
\Prob[\fkNN_{k,n}(x)\ne f^*(x) ] 
& \ge \Prob[\fkNN_{k,n}(x)\ne f^*(x) \mid R_{(k+1)}(x) \in \calR_b]  - \Prob[R_{(k+1)}(x) \not\in \calR_b]\nonumber\\
&\ge \frac{1}{2}- \Prob[R_{(k+1)}(x) \not\in \calR_b].\label{eq:tk-fknn} 
\end{align}

Note that the distribution of the signal is bounded with $\Norm{\Pi(X)}_2\le 5 \triangleq R $.
We keep $R$ in the expression to help understand the proof, but note that $R$ in our setup is an absolute constant.
Next, 
we prove that, for a small constant $\tau >0$, 
\begin{equation}
\label{eq:bad-subset}
I \triangleq \qth{\sqrt{\mu+4R^2- \tau\sigma^2}, \sqrt{\mu+\tau\sigma^2}}
\subseteq \calR_b,
\end{equation}
where 
\begin{align*}
    &\mu \triangleq\Expect[\Norm{\Pi_\perp(X-x)}_2^2]=d-2+ \pth{\Norm{x}_2^2- (x^{(1)})^2 - (x^{(2)})^2},\\
    &\sigma^2 \triangleq\var[\Norm{\Pi_\perp(X-x)}_2^2]=2(d-2)+4 \pth{\Norm{x}_2^2- (x^{(1)})^2 - (x^{(2)})^2}.
\end{align*}

Similar to the proof of~\prettyref{prop:logit-dominance}, we get
\[
\mu_{1,r}(x)-\mu_{0,r}(x)
    = \int f(x,r^2-u^2) \cdot (\nu_1(x,u) - \nu_0(x,u)) \cdot 2u \diff u. 
\]
Since the distribution of the signal is bounded by $R$, we have $\nu_i(x, t) = 0$ when $t\le 0$ and $\nu_i(x, t) =1$ when $t\ge 2R$. 
Then, 
\begin{align}
&\mu_{1,r}(x)-\mu_{0,r}(x)\nonumber\\
=~& f(x,r^2) \int (\nu_1(x,u) - \nu_0(x,u)) \cdot 2u \diff u 
+ \int (f(x,r^2-u^2)-f(x,r^2)) \cdot (\nu_1(x,u) - \nu_0(x,u)) \cdot 2u \diff u\nonumber\\
\le~& f(x,r^2) \int (\nu_1(x,u) - \nu_0(x,u)) 2u \diff u
+ \int_0^{2R} |f(x,r^2-u^2)-f(x,r^2)| \cdot 2u \diff u. \label{eq:ub-mu-diff}
\end{align}

\paragraph{First term of~\eqref{eq:ub-mu-diff}.}
For the nonnegative random variable $\Norm{\Pi(X-x)}_2^2$,
\begin{align*}
\Expect[\Norm{\Pi(X-x)}_2^2 | Y=i] 
&= \int_0^\infty \Prob[\Norm{\Pi(X-x)}_2^2>t | Y=i] \diff t \\
&= \int_0^\infty (1-\nu_i(x, u)) 2u \diff u,
\end{align*}
where $i \in\{0,1\}$. 
Then, applying~\eqref{eq:moment-diff} yields that
\begin{equation}
\label{eq:shift-dominance}
\int (\nu_1(x,u) - \nu_0(x,u)) 2u \diff u
= \int_0^\infty \pth{ (1-\nu_0(x, u) ) - (1-\nu_1(x, u) ) }  2u \diff u
\le -0.1.
\end{equation}

\paragraph{Second term of~\eqref{eq:ub-mu-diff}.}
We define a shifted and rescaled function of $f(x,u)$ as 
\[
h(z) \triangleq \sigma f(x,\mu + \sigma z).
\]
For $r\in I$, let $z=\frac{r^2-\mu}{\sigma}$.
Then we get
\begin{align*}
\int_0^{2R} |f(x,r^2-u^2)-f(x,r^2)| \cdot 2u \diff u
&=\frac{1}{\sigma} \int_0^{4R^2} \abs{h\pth{z-\frac{u}{\sigma}}-h(z)} \diff u\\
&\le \frac{1}{\sigma} \int_0^{4R^2} \int_{0}^{u/\sigma} |h'(z-v)| \diff v \diff u.
\end{align*}

\begin{lemma}
\label{lmm:noncentral}
    There exists an absolute constant $C$ such that,  for any $\tau$ satisfying $d^{-1/3} \lesssim \tau \le \frac{1}{8}$,
    \begin{align*}
        \max_{|t| \le \tau \sigma} \frac{|h'(t)|}{h(t)} \le C\tau\sigma.
    \end{align*}
\end{lemma}

By Lemma~\ref{lmm:noncentral}, there exists a constant $C>0$ such that $|h'(t)|\le C \tau \sigma  h(t)$ for $|t| \le \tau \sigma$.
For any $0< v <4R^2/\sigma$, it follows from the range of $I$ in~\eqref{eq:bad-subset} that $-\sigma\tau \le z-v \le z \le  \sigma \tau$. 
Applying Grönwall's inequality yields  
\[
|h'(z-v)| \le C \tau \sigma h(z-v)
\le C \tau \sigma e^{ C \sigma v}h(z).
\]
Consequently,
\begin{align}
\int_0^{2R} |f(x,r^2-u^2)-f(x,r^2)| \cdot 2u \diff u 
&\le \frac{1}{\sigma} \int_0^{4R^2} \int_{0}^{u/\sigma} C \tau \sigma e^{ C \sigma v}h(z) \diff v \diff u \nonumber \\
&= \frac{\tau h(z)}{\sigma} \int_0^{4R^2} (e^{Cu} -1) \diff u \nonumber \\
&= \tau f(x,r^2) \frac{e^{4 C R^2} - 1 - 4 C R^2}{C}. \label{eq:gronwall}
\end{align}

Therefore, by choosing $\tau< \frac{0.1 C}{e^{4 C R^2} - 1 - 4 C R^2}$, we have $\tau f(x,r^2) \frac{e^{4 C R^2} - 1 - 4 C R^2}{C}< 0.1 f(x,r^2)$.
Plugging \eqref{eq:shift-dominance} and \eqref{eq:gronwall} in \eqref{eq:ub-mu-diff}, we get 
\[
\mu_{1,r}(x)-\mu_{0,r}(x)<0,
\]
and thus $r\in \calR_b$.
We completed the proof of $I \subseteq \calR_b $ as stated in~\eqref{eq:bad-subset}.

It remains to upper bound $\Prob[R_{(k+1)}(x) \not\in \calR_b]$ in \eqref{eq:tk-fknn}.
If the event $R_{(k+1)}(x) \in \calR_b^c \subseteq I^c$ occurs, then there must exist a training data point with $\Norm{X_i-x}_2 \in I^c$.
By the union bound,
\begin{align*}
    &\Prob[R_{(k+1)}(x) \not\in \calR_b]\\
    \le~& n \Prob[\Norm{X-x}_2 \in I^c]\\
    =~& n \pth{ \Prob[\Norm{X-x}_2^2 \ge \mu +\tau \sigma^2] 
    + \Prob[\Norm{X-x}_2^2 \le \mu + 4R^2 - \tau \sigma^2]  }.
\end{align*}
Since $\Norm{\Pi(X-x)}_2^2 \le 4R^2$, we have
\begin{align*}
& \prob{\Norm{X-x}_2^2 \ge \mu + \tau \sigma^2 } 
\le \prob{  \Norm{\Pi_\perp(X-x)}_2^2 \ge \mu + \tau \sigma^2 - 4R^2},  \\
& \prob{\Norm{X-x}_2^2 \le \mu + 4R^2-\tau \sigma^2 } 
\le \prob{  \Norm{\Pi_\perp(X-x)}_2^2 \le \mu - (\tau \sigma^2 - 4R^2) }.
\end{align*}
By the data model $X^{(i)} \sim \calN(0,1)$, we apply the Bernstein inequality to upper bound the probability of the above two tail events. 
To this end, the sub-exponential norm $\Norm{\cdot }_{\psi_1}$of $\Norm{\Pi_\perp(X-x)}^2- \mu = \sum_{i=3}^{d} (X^{(i)}-x^{(i)})^2-\Expect(X^{(i)}-x^{(i)})^2$ is upper bounded as following:
\begin{align*}
\Norm{(X^{(i)}-x^{(i)})^2 - \Expect (X^{(i)}-x^{(i)})^2}_{\psi_1}
\lesssim \Norm{(X^{(i)}-x^{(i)})^2}_{\psi_1}
= \Norm{X^{(i)}-x^{(i)}}_{\psi_2}^2
\lesssim 1 + x_i^2.
\end{align*}
Therefore, since $\tau \sigma^2 - 4R^2\ge \frac{\tau \sigma^2}{2} \gtrsim \tau d$,  applying Bernstein inequality (see, e.g., \cite[Theorem 2.8.1]{Vershynin2018}) yields that
\begin{align*}
\prob{\abs{\Norm{\Pi_\perp(X-x)}^2 - \mu } \ge \tau \sigma^2 - 4R^2}
& \le 2\exp\pth{-c' \frac{\tau^2 d^2}{\sum_{i=3}^{d}(1+x_i^2)^2} \wedge \frac{\tau d}{ \max_{i\in[n]}(1+x_i^2)} } \\
& \le 2\exp\pth{-c \frac{\tau^2 d}{\beta }\wedge \frac{\tau d}{\beta'}  },
\end{align*}
for absolute constants $c$ and $c'$.
We conclude that
\[
\Prob[R_{(k+1)}(x) \not\in \calR_b]
\le 2n \exp\pth{-c \frac{\tau^2 d}{\beta }\wedge \frac{\tau d}{\beta'}  }.
\]
The proof is completed by plugging the above in \eqref{eq:tk-fknn}.
\end{proof}

\begin{proof}[Proof of \prettyref{lmm:noncentral}]
Let $\lambda = \Norm{\Pi_\perp(x)}_2^2$, and $g$ denote the probability density function of the non-central $\chi^2$ distribution of degree $d_\sfn=d-2$ and non-centrality $\lambda$.
Then, we obtain from \cite[Corollary 1.3.5]{Muirhead2009} that
\[
\mu = \Expect[\Norm{\Pi_\perp(X-x)}_2^2]=d_\sfn+\lambda,
\quad
\sigma^2 = \var[\Norm{\Pi_\perp(X-x)}_2^2] = 2(d_\sfn+2\lambda).
\]
By a change of variables, the standardized density function $h$ can be represented in terms of $g$ as $h(t)=\sigma g(\mu+ \sigma t)$, and the desired result is equivalent to
\begin{equation}\label{eq:goal-non-central-chi2}
\max_{t\in [\mu-\tau \sigma^2, \mu+\tau \sigma^2]} \frac{|g'(t)|}{g(t)}
\le C\tau.
\end{equation}

We prove the result for even degrees $d_\sfn$ in which case $m\triangleq \frac{d_\sfn}{2}-1$ is an integer. The case for odd $d_\sfn$ follows from analogous arguments. 
In the even-degree case, $g$ permits the following analytic expression by \cite[Theorem 1.3.4]{Muirhead2009} and \cite[Eq.~9.6.47]{AS64}
\begin{align*}
    g(t) = \frac{1}{2}e^{-(t+\lambda)/2} \pth{\frac{t}{\lambda}}^{m/2} I_{m} (\sqrt{\lambda t}),
\end{align*}
where $I_m$ is the modified Bessel function of the first kind defined as
\begin{equation}\label{eq:Im-series}
I_m(z) = \pth{\frac{z}{2}}^m \sum_{j=0}^\infty \frac{(z^2/4)^j}{\Gamma(m+j+1)j!}.
\end{equation}
By taking the derivative of $g$ and applying the recurrent rule of $I_m$ \cite[Eq.~9.6.26]{AS64}, we get,
\begin{align}
    g'(t) = \frac{1}{4}e^{-(t+\lambda)/2} \pth{\frac{t}{\lambda}}^{m/2} \pth{-I_m(\sqrt{\lambda t}) + \sqrt{\frac{\lambda}{t}}I_{m-1}(\sqrt{\lambda t})}.
\end{align}
By directly replacing $I_m$ by its series sum expression~\eqref{eq:Im-series}, we get 
\begin{align}
    \frac{g'(t)}{g(t)} 
    & = \frac{-1}{2}\pth{1- \sqrt{\frac{\lambda}{t}} \frac{I_{m-1}(\sqrt{\lambda t})}{I_{m}(\sqrt{\lambda t})}} \nonumber \\
    & = \frac{-1}{2}\pth{1- \frac{2}{t}  \frac{\sum_{j =0 }^\infty a_j (m+j)}{\sum_{j =0 }^\infty a_j} } \nonumber \\
    & = \frac{-1}{2}\pth{1 - \frac{2 m + \lambda }{t}} + \frac{\sum_{j =0 }^\infty (j - \lambda / 2) a_j}{ t \sum_{j =0 }^\infty a_j},\label{eq:gdot-g} 
\end{align}
where
\begin{align*}
    a_j = \frac{ (\lambda t / 4)^j}{ j ! \ (m+j)!}, 
    \qquad
    \frac{a_j}{a_{j-1}} = \frac{\lambda t / 4}{j (m+j)}.
\end{align*}

Note that $ 2\mu\le \sigma^2 \le 4\mu$ and $2m+\lambda = \mu-2$.
When $|t-\mu| \le \tau \sigma^2$ and $ \frac{2}{\sigma^2} \le \tau \le \frac{1}{8}$, we have
\[
\frac{1}{2}\abs{1 - \frac{2 m + \lambda }{t}} 
\le \frac{\tau \sigma^2 + 2}{2(\mu-\tau \sigma^2)}
\le 8 \tau.
\]
To bound the second term on the right-hand side of~\eqref{eq:gdot-g}, we split the infinite sum into three parts:
\begin{align*}
    I_1 &= \mathbb{Z}_{+} \cap [0, (1-4\tau)\lambda/2], \\
    I_2 &=  \mathbb{Z}_{+}\cap [(1-4\tau)\lambda/2, (1+4\tau)\lambda/2], \\ 
    I_3 &= \mathbb{Z}_{+} \cap [(1 + 4\tau)\lambda/2, \infty).
\end{align*}

For the summation over $I_2$, we have
\begin{align*}
    \sum_{j \in I_2} |j - \lambda /2| a_j
    \le 2\tau \lambda  \sum_{j \in I_2} a_j 
    \le 2\tau \lambda  \sum_{j=0}^\infty a_j.
\end{align*}
Since $\tau \le 1/8$ and $t\ge \mu-\tau \sigma^2 \ge (1-4\tau)\lambda$,
\[
\frac{\sum_{j \in I_2} |j - \lambda /2|a_j  }{t \sum_{j=0}^\infty a_j}
\le \frac{2\tau \lambda}{t}
\le \frac{2\tau}{1-4\tau}
\le 4\tau.
\]


Next, for $j \in I_3$, applying $t\le \mu + \tau \sigma^2 = (1 + 2\tau)d_\sfn + (1 + 4\tau)\lambda$ yields
\begin{align*}
    \frac{a_j}{a_{j-1}} 
    = \frac{\lambda / 2}{j} \frac{t / 2}{(m+j)}
    &\le \frac{1}{1 + 4 \tau} \frac{(1 + 2\tau)d_\sfn + (1 + 4\tau)\lambda}{d_{\sfn}-2 + (1 + 4\tau)\lambda} \\
    &\le \frac{1 + 3\tau}{1 + 4 \tau} \le \frac{1}{1 + \frac{\tau}{2}},
\end{align*}
where the first inequality applied the definition of $m$ and the range of $j\in I_3$, the second inequality holds since $(1 + 2\tau)d_\sfn\le (1+3\tau)(d_\sfn-2)$ when $d_\sfn\tau \ge 4$, and the last inequality holds since $\tau \le \frac{1}{8}$.
Therefore, we have for $j \in I_3$,
\begin{align*}
    a_{j} \le \frac{a_{\lceil (1 + 4\tau)\lambda/2 \rceil}}{(1+\frac{\tau}{2})^{j - \lceil (1+4\tau)\lambda/2 \rceil }}.
\end{align*}
Consequently,
\begin{align*}
    \sum_{j \in I_3} |j - \lambda /2| a_j 
    &= \sum_{j=0}^\infty (j+\Ceil{(1+4\tau)\lambda/2}-\lambda/2)  a_{j+\Ceil{(1+4\tau)\lambda/2}}  \\
    &\le \sum_{j=0}^\infty j a_{j+\ceil{(1 + 4\tau)\lambda/2}} + 2\tau \lambda \sum_{j \in I_3} a_j \\
    &\le a_{\lceil (1 + 4\tau)\lambda/2 \rceil} \sum_{j=0}^\infty \frac{j }{(1+\frac{\tau}{2})^j} +  2\tau \lambda\sum_{j \in I_3} a_j\\
    & \lesssim \pth{\frac{1}{\tau^2} + \tau \lambda} \sum_{j = 0 }^{\infty} a_j.
\end{align*}
Since $\tau \gtrsim d_{\sfn}^{-1/3}$ and thus $\tau \lambda \gtrsim \tau^{-2}$, we have
\[
\frac{\sum_{j \in I_3} |j - \lambda /2| a_j }{t \sum_{j = 0 }^{\infty} a_j}
\lesssim \tau.
\]


Finally, we can similarly upper bound the term in $I_1 $: for $\tau\le \frac{1}{8}$,
\begin{align*}
    \frac{a_j}{a_{j-1}} 
    \ge \frac{\lambda / 2}{j} \frac{t / 2}{(m+j)} 
    \ge \frac{1}{1 - 4 \tau} \frac{(1-2\tau)d_\sfn + (1-4\tau)\lambda}{d_{\sfn}-2 + (1-4\tau)\lambda} \ge 1 + \tau.
\end{align*}
The remaining analysis is similar to that for $I_3$.
The proof is completed. 
\end{proof}


\end{document}